\documentclass{article}
\usepackage{arxiv}
\usepackage[utf8]{inputenc} 
\usepackage[T1]{fontenc}    
\usepackage{hyperref}       
\usepackage{url}            
\usepackage{booktabs}       
\usepackage{amsfonts}       
\usepackage{nicefrac}       
\usepackage{microtype}      
\usepackage{lipsum}
\usepackage{graphicx}
\graphicspath{ {./images/} }
\usepackage{graphicx}%
\usepackage{multirow}%
\usepackage{amsmath,amssymb,amsfonts}%
\usepackage{amsthm}%
\usepackage{mathrsfs}%
\usepackage[title]{appendix}%
\usepackage{textcomp}%
\usepackage{manyfoot}%
\usepackage{booktabs}%
\usepackage{algorithmicx}%
\usepackage{listings}%
\usepackage{algorithm}
\usepackage{algpseudocode}
\usepackage{researchpack}
\usepackage{adjustbox}
\usepackage{enumitem}
\usepackage{hyperref}
\usepackage[dvipsnames,table]{xcolor}
\usepackage[numbers,sort&compress]{natbib}
\usepackage[capitalise,nameinlink]{cleveref}

\usepackage{relsize}
\usepackage{tikz}
\usetikzlibrary{shapes,decorations,arrows,calc,arrows.meta,fit,positioning}

\hypersetup{
    colorlinks=true,
    citecolor=blue,
    linkcolor=blue,
    urlcolor=blue,
}
\graphicspath{images/}

\tikzset{
    -Latex,auto,node distance =1 cm and 1 cm,semithick,
    state/.style ={ellipse, draw, minimum width = 0.7 cm},
    point/.style = {circle, draw, inner sep=0.04cm,fill,node contents={}},
    bidirected/.style={Latex-Latex,dashed},
    el/.style = {inner sep=2pt, align=left, sloped}
}

\theoremstyle{thmstyleone}%
\theoremstyle{thmstyletwo}%
\theoremstyle{thmstylethree}%

\newcommand{\FY}{\ensuremath{F_1(Y)}\xspace}

\newcommand{\CAUC}{\ensuremath{\mathsf{AUC}(C)}\xspace}
\newcommand{\Disent}{\ensuremath{\mathsf{DIS}}\xspace}
\newcommand{\Leakage}{\ensuremath{\mathsf{LEAK}}\xspace}
\newcommand{\OIS}{\ensuremath{\mathsf{OIS}}\xspace}
\newcommand{\MacroRecall}{\ensuremath{\mathrm{MRc}}\xspace}

\newcommand{\MacroPrecision}{\ensuremath{\mathrm{MPr}}\xspace}

\newcommand{\SHAPES}{\texttt{Shapes3d}\xspace}
\newcommand{\CelebA}{\texttt{CelebA}\xspace}
\newcommand{\CUB}{\texttt{CUB}\xspace}

\newcommand{\CBM}{\texttt{CBM}\xspace}

\newcommand{\PHCBM}{\texttt{PH-CBM}\xspace}
\newcommand{\LABO}{\texttt{LaBo}\xspace}
\newcommand{\LFCBM}{\texttt{LF-CBM}\xspace}
\newcommand{\VLGCBM}{\texttt{VLG-CBM}\xspace}
\newcommand{\DNCBM}{\texttt{DN-CBM}\xspace}

\newcommand{\method}{\texttt{Na\"ive}\xspace}
\newcommand{\SAGA}{\texttt{GLM-SAGA}\xspace}

\usepackage{pifont}
\newcommand{\cmark}{\textcolor{black}{\ding{51}}}%
\newcommand{\xmark}{\textcolor{black}{\ding{55}}}%
\newcommand{\ccmark}{\textcolor{black}{\ding{51}}}%
\newcommand{\xxmark}{\textcolor{black}{\ding{55}}}%

\title{If Concept Bottlenecks are the Question, are Foundation Models the Answer?}

\author{
 Nicola Debole \\
  DISI, University of Trento, Italy\\
  \texttt{nicola.debole@unitn.it} \\
   \And
 Pietro Barbiero \\
  IBM Research Zurich, Switzerland\\
  \texttt{pietro.barbiero@ibm.com} \\
  \And
 Francesco Giannini \\
  Faculty of Sciences, \\
  Scuola Normale Superiore, Italy\\
  \texttt{francesco.giannini@sns.it} \\
  \And
 Andrea Passerini \\
  DISI, University of Trento, Italy\\
  \texttt{andrea.passerini@unitn.it} \\
  \And
 Stefano Teso\footnotemark[2] \\
  CIMeC, University of Trento, Italy\\
  DISI, University of Trento, Italy\\
  \texttt{stefano.teso@unitn.it} \\
  \And
 Emanuele Marconato\thanks{Corresponding author.}\hspace{0.2em} \thanks{Shared last author.} \\
  DISI, University of Trento, Italy\\
  \texttt{emanuele.marconato@unitn.it} \\
}

\begin{document}
\maketitle
\begin{abstract}
Concept Bottleneck Models (CBMs) are neural networks designed to conjoin high performance with \textit{ante-hoc} interpretability.
    CBMs work by first mapping inputs (\eg images) to high-level concepts (\eg
    visible objects and their properties) and then use these to solve a downstream task (\eg tagging or scoring an image) in an interpretable manner.
    Their performance and interpretability, however, \textit{hinge on the quality of the concepts they learn}.
    The go-to strategy for ensuring good quality concepts is to leverage expert annotations, which are expensive to collect and seldom available in applications.
    Researchers have recently addressed this issue by introducing ``VLM-CBM'' architectures that replace manual annotations with weak supervision from foundation models.
    It is however unclear what is the impact of doing so on the quality of the learned concepts.
    To answer this question, we put state-of-the-art VLM-CBMs to the test, analyzing their learned concepts empirically using a selection of significant metrics.
    Our results show that, depending on the task, VLM supervision can sensibly differ from expert annotations, and that concept accuracy and quality are not strongly correlated.
    Our code is available at \href{https://github.com/debryu/CQA}{https://github.com/debryu/CQA}.
\end{abstract}


\section{Introduction}

Concept Bottleneck Models (CBMs) \citep{koh2020concept} are a popular class of neural networks that aim to resolve the traditional trade-off between interpretability and accuracy.
In a nutshell, a CBM comprises two learnable modules: a \textit{concept extractor} and an \textit{inference layer}.
The former maps the input into an activation vector of high-level concepts, while the latter is generally a white-box model, typically a (sparse) linear layer, that computes predictions from the extracted concepts.
For instance, given the image of a red ball, a CBM would encode it into a vector of concept activations -- typically logits -- for the events $(\textit{``shape''} = {\tt sphere})$, $(\textit{``color''} = {\tt red})$, and all other combinations established by a pre-defined vocabulary.
This vector plays the role of a \textit{\textbf{bottleneck}} encoding interpretable sufficient statistics for the final prediction.
CBMs are \textit{explainable-by-design} in the sense that they allow tracing any decision back to the concepts most relevant for it.
At the same time, they support representation learning and can achieve high performance for complex data (\eg images) that are beyond the reach of traditional white-box models.
Their modular architecture also brings other benefits, such as support for \textit{correcting} the model's predictions via user interventions \citep{koh2020concept,zarlenga2024learning,shin2023closer,steinmann2024learning} and for \textit{debugging} the model itself using explanation-based feedback \citep{stammer2021right,bontempelli2023concept}.
For these reasons, the same two-step setup has been instantiated into a number of recent classifiers \citep{sawada2022csenn, sawada2022concept,barbiero2023interpretable, marconato2022glancenets,kim2023probabilistic,vandenhirtz2024stochastic} and generative models \citep{ismail2023concept,dominici2024counterfactual}.

It is however possible that the \textit{\textbf{learned concepts may not match the semantics that humans attribute to them}}.
The more they deviate, the more opaque the model's decision process becomes, compromising all benefits that CBMs are designed to provide, including interpretability and intervenability \citep{mahinpei2021promises, marconato2023interpretability, furby2023towards, raman2023concept, lai2024faithful}.

\begin{figure}[!t]
    \centering
    \includegraphics[width=0.98\textwidth]{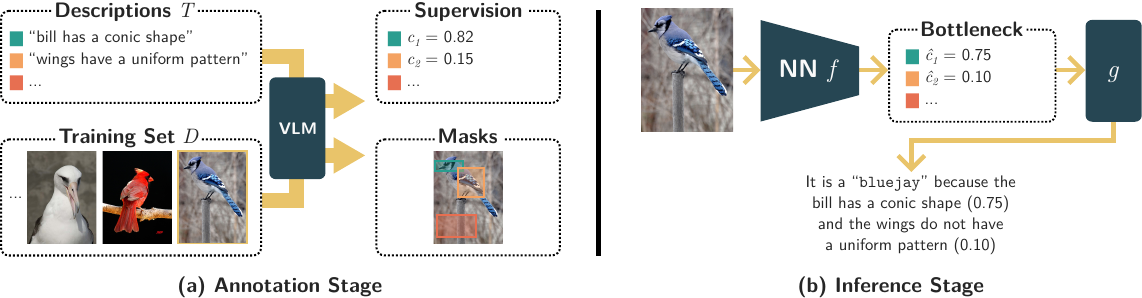}
    \caption{
    \textbf{A prototypical VLM-CBM}.
    \textbf{Left}: given textual descriptions $\calT$ of visual concepts, a VLM is used to annotate a training set $\calD$ with per-concept labels or activation scores and (optionally) masks indicating where each concept activates.
    \textbf{Right}: the supervision is typically used to fine-tune  a backbone $f$ that extracts concepts from new inputs and a linear layer $g$ for inferring predictions from the concepts.
    See \cref{sec:prelims} for architecture-specific differences.
    }
    \label{fig:cbms}
\end{figure}

The go-to strategy for aligning concepts to human semantics is to exploit manual concept annotations, however these are expensive to collect and thus often unavailable in applications.
An increasingly prominent solution is to replace manual annotations with \textit{weak supervision} obtained by querying Vision-Language Models (VLMs) with natural language descriptions of the desired concepts \citep{oikarinen2023label, yang2023language, srivastava2024vlgcbm, rao2024discover}.
In essence, this procedure distills the concepts learned by a source VLM into a target CBM (cf. \Cref{fig:cbms}), with the explicit goal of extending the reach of CBMs to tasks in which no manual annotations are available.
The resulting \textit{\textbf{VLM-CBMs}} offer competitive prediction performance and in many cases admit fast inference.\footnote{This does not apply to VLM-CBMs that invoke the VLM during inference, see \cref{sec:prelims}.}

VLMs, however, may be no silver bullet.  In fact, although trained on very large corpora, they can suffer from hallucinations \citep{huang2023survey}, shortcuts \citep{yuan2024llms}, and subpar logical \citep{calanzone2025logically} and conceptual \citep{sahu2022unpacking} consistency, and the distillation step risks to pass these defects onto the target CBM.
Moreover, even highly accurate concepts can capture spurious contextual information -- due to, \eg \textit{entanglement} \citep{higgins2018towards, scholkopf2021toward, suter2019robustly, eastwood2018framework} and \textit{leakage} \citep{margeloiu2021concept, mahinpei2021promises, marconato2022glancenets, havasi2022addressing}, illustrated in \Cref{fig:concept-issues} -- in which case the model's explanations become misleading.
Unfortunately, most works on VLM-CBMs are chiefly concerned with downstream accuracy \citep{oikarinen2023label, yang2023language, srivastava2024vlgcbm, rao2024discover, yuksekgonul2023post} rather than quality of the learned concepts.
As a result, it is still unclear whether VLM-CBMs learn concepts of high enough quality.

We fill this gap by directly assessing the concepts learned by representative VLM-CBM architectures -- namely, language-in-a-bottle \citep{yang2023language}, label-free CBMs \citep{oikarinen2023label}, and vision-language-guided CBMs \citep{srivastava2024vlgcbm}; see \cref{tab:cbms} for a breakdown of different architectures -- using a variety of metrics focusing on diverse, relevant aspects of concept quality.
Our experiments on three datasets of increasing complexity indicate that:
(\textit{i}) \textit{VLM concept supervision can sensibly differ from manual annotations}, even when textual descriptions $\mathcal{T}$ of concepts are provided by experts;
(\textit{ii}) in many cases \textit{VLM-CBMs can output correct label predictions by leveraging low-quality concepts} and, more generally, prediction performance and concept quality are not strongly correlated.
Overall, our results show that, while VLM-CBMs enjoy wider applicability compared to regular CBMs, \textit{\textbf{replacing manual with VLM supervision can come at a substantial cost for concept quality}}, and thus interpretability.
With this work, we hope to prompt more stringent evaluation of concepts learned by state-of-the-art CBMs and to refocus the design of VLM-CBMs on robustness to annotation mistakes and interpretability.

\section{The Family of Concept Bottleneck Models}
\label{sec:prelims}

Concept Bottleneck Models (CBMs) \citep{koh2020concept} combine two neural modules: a \textit{concept extractor} $f:\calX \to \bbR^k$ mapping an input $\vx \in \calX$ to a concepts activation vector $\hat \vc \in \bbR^k$, which encodes logits of $k$ high-level concepts describing the input; and a white-box \textit{inference layer} $g: \bbR^k \to \calY$ mapping the concept activations to a prediction $\hat y \in \calY$.
Combining $f$ and $g$ yields the complete predictor.
Despite not being restricted to specific data or tasks, CBMs are often used for image classification.  This is the use case we focus on.

While the concept extractor is an expressive (but black-box) neural backbone -- enabling CBMs to achieve competitive \textit{label} accuracy -- the inference layer is white-box.
This means that CBMs can naturally explain their own inferences, in that it is easy to identify what concepts contributed the most to each prediction.
For instance, if the inference step is implemented -- as customary -- with a sparse linear layer, then a CBM \textit{explanation} elucidates to what extent each concept activation $\hat c_j$ contributes to a prediction $\hat y$ by considering the weights $w_{\hat y, j}$ of the linear layer \citep{kim2018interpretability}.
See \Cref{fig:cbms} (right) for an example.
A major benefit of this setup is \textit{intervenability}:  whenever concept are mis-predicted, users can readily replace them with better estimates \citep{koh2020concept,shin2023closer,zarlenga2024learning,chauhan2023interactive}.  It is also possible to teach CBMs to reapply past interventions to unseen instances \citep{steinmann2024learning} or to learn when interventions are required at training time \citep{espinosa2023learning}, thus implementing a form of interactive debugging \citep{teso2023leveraging}.
CBM architectures differ in how they define, extract and use their concepts, yet popular variants like Concept Embedding Models \citep{zarlenga2022concept} and GlanceNets \citep{marconato2022glancenets} all follow the same modular setup.
We leave an in-depth discussion of other concept-based models to \cref{sec:related-work}.

\paragraph{Learning with expert supervision}
CBM explanations are interpretable only as long as the concepts are \textit{aligned} with the semantics that humans associate to them \citep{marconato2023interpretability, fokkema2025sample}.
Most CBMs exploit dense concept-level annotations during training to ensure that this is the case.  They typically do so by combining two cross-entropy losses, one for the labels $\calL_Y = \bbE_{(\vx, y)} \big[ - \log p ( y \mid \vx) \big]$ and one for the concepts $\calL_{\vC} = \bbE_{(\vx, \vc)} \big[ - {\textstyle \frac{1}{k} \sum_{j \in [k]}} \log p(c_j \mid \vx) \big]$, although the exact formulations vary between architectures \citep{koh2020concept, zarlenga2022concept, marconato2022glancenets}.
Here, $p(c_j \mid \vx) = \mathrm{sigmoid}(f(\vx)_j)$ is the predictive distribution of the $j$-th concept and the expectations run over the training data.
The two modules can be trained either independently, sequentially or jointly \citep{koh2020concept}, and the choice can impact the learned representations \citep{mahinpei2021promises}.  Most architectures discussed below follow a sequential setup: the concept extractor is learned first, by minimizing $\calL_{\vC}$, and it is then used to train the inference layer via $\calL_Y$.  Note also that, in most complex tasks, the concept extractor is pretrained and fine-tuned, rather than trained from scratch.

\begin{table}[!t]
    \centering
    \caption{\textbf{Comparison of CBMs and VLM-CBMs}.
    The columns which \textsc{VLM} is used,
    which kind of concept supervision (\textsc{Concept Sup.}) is required,
    whether the textual concept descriptions (that is, $\calT$) can be supplied externally,
    and whether the concept extractor requires a pretrained model (\textsc{Pretrained})
    }
    \label{tab:cbms}
    \scriptsize
    \begin{tabular}{lcccc}
        \toprule
        {\sc Model}
            & {\sc VLM}
            & {\sc Concept Sup.}
            & {\sc External $\calT$}
            & {\sc Pretrained}
        \\
        \midrule
        \rowcolor[HTML]{EFEFEF}
        \CBM \citep{koh2020concept}
            & --
            & Labels
            & \cmark$^*$
            & \xxmark$^{**}$ 
        \\
        \LABO \citep{yang2023language}
            & {\tt CLIP} \citep{radford2021learning}
            & Similarity
            & \cmark \ \ 
            & \ccmark \ \ \
        \\
        \rowcolor[HTML]{EFEFEF}
        \LFCBM \citep{oikarinen2023label}
            & {\tt CLIP} \citep{radford2021learning}
            & Similarity
            & \cmark \ \ 
            & \xxmark \ \ \
        \\
        \VLGCBM \citep{srivastava2024vlgcbm}
            & {\tt G-DINO} \citep{liu2024grounding}
            & Labels, Bboxes
            & \cmark \ \ 
            & \xxmark \ \ \
        \\
        \rowcolor[HTML]{EFEFEF}
        \PHCBM \citep{yuksekgonul2023post}
            & {\tt CLIP} \citep{radford2021learning} 
            & Similarity
            & \cmark \ \ 
            & \ccmark \ \ \
        \\
        \DNCBM \citep{rao2024discover}
            & {\tt CLIP} \citep{radford2021learning} 
            & Similarity
            & \xmark \ \ 
            & \ccmark \ \ \
        \\
        \rowcolor[HTML]{EFEFEF}
        \method        
            & {\tt LLaVa} \citep{2023xtuner}
            & Labels
            & \cmark \ \ 
            & \xxmark \ \ \
        \\
        \bottomrule
    \end{tabular}
    
    \footnotesize
    
    $^*$ The concept vocabulary can be customized in standard CBM too, but this comes at the cost of having to collect manual annotations for any new concepts. \\ 
    $^{**}$ Some variants of CBMs \citep{dominici2024anycbmsturnblackbox, laguna2024beyond,sawada2022concept} require using pretrained modules.
\end{table}

\paragraph{Learning with VLM supervision}

A key issue of CBMs  is that concept annotations are \textit{not} available in most applications.  In these cases, $\calL_{\vC}$ cannot be used during learning, and weak supervision coming from the label annotations via $\calL_{\vY}$ can be insufficient to prevent concepts from deviating, even radically, from the ground-truth ones \citep{bortolotti2025shortcuts}.
In an attempt to widen their applicability, researchers have recently proposed several ``VLM-CBMs'' that forego expert annotations in favor of a pre-trained Vision-Language Model (VLM).
The idea is that, when asked to detect a set of preselected concepts, identified by textual descriptions $\calT = \{t_1, \ldots, t_k\}$, in any image, a sufficiently large VLM trained on a wealth of diverse data will likely produce a high-quality response.
Next, we introduce the key steps involved in learning a VLM-CBM (\Cref{fig:cbms}).\\

\noindent
\textit{\textbf{Step 1: Obtaining concept supervision from the VLM.}}
A popular option is to obtain concept activations by matching textual concept descriptions and training images in the VLM's \textit{embedding space}.
For instance, Label-free CBMs (\LFCBM{s}) \citep{oikarinen2023label}, Language-in-a-Bottle (\LABO) \citep{yang2023language} and Post-hoc CBMs (\PHCBM) \citep{yuksekgonul2023post} use CLIP's visual encoder $E_X$ and text encoder $E_T$ \citep{radford2021learning} to compute (variants of) the cosine similarity $c_j = \mathrm{sim}(E_X(\vx), E_T(t_j))$ between image $\vx$ and text $t_j$.\footnote{We use the same notation for both scores and labels, for simplicity.}
A more direct route is to collect binary annotations $c_j$ by \textit{prompting} the VLM: one simply feeds a training input $\vx$ to the VLM and asks whether a concept described by $t_j$ is present.
Vision-Language-Guided CBMs (\VLGCBM{s}) \citep{srivastava2024vlgcbm} do so with Grounding Dino \citep{liu2024grounding}.  As a bonus, they also collect per-concept bounding-boxes and use them to filter out low-quality hits.
In either case, one obtains similarity scores or binary labels $\vc \in \bbR^k$ for each data point.

Finally, Discover-then-Name (\DNCBM) \citep{rao2024discover} stands out in that it extracts concept activations using sparse autoencoders. Since the extracted concepts are anonymous, they are assigned a name by decoding the image representation to text using CLIP.  One downside is that this makes it impossible to prespecify the vocabulary $\calT$.
\\

\noindent
\textit{\textbf{Step 2: Training the concept extractor with VLM supervision.
}} 
\LFCBM updates the backbone using a modified concept loss $\calL_{\vC}$ that regresses on the similarity scores, while \VLGCBM uses the concept annotations to power the cross-entropy loss $\calL_{\vC}$.  It also optionally augments the training set by adding cut-outs of the input image identified by the concepts' bounding boxes, so as to encourage learned concepts to be independent of their background.
\LABO and \PHCBM are special as they have no backbone\footnote{When viewed as VLM-CBMs, they are in fact equivalent.} and apply the inference layer directly to the similarity scores.\footnote{The downside is the non-negligible runtime cost of running CLIP at inference time.}
\\

\noindent
\textit{\textbf{Step 3: Training the top linear layer.}}  Once the concepts have been learned, the top layer is trained by optimizing $\calL_{\vY}$.  The size of concept explanations is kept under control by incorporating a sparsity penalty -- such as $\ell_1$-regularization or group lasso \citep{oikarinen2023label, srivastava2024vlgcbm} -- or by normalizing the weights \citep{yang2023language}.


\paragraph{Issues with VLM-CBMs}  VLM-CBMs extend the reach of CBMs to tasks that lack dense annotations, yet they are not devoid of risks in that their concepts may not always be high quality.
For instance, some concepts learned by popular VLM-CBMs can be irrelevant and/or non-visual (cf. \cref{tab:invalid_concepts}) and still affect the model's decision making \citep{srivastava2024vlgcbm, yang2023language}.
The VLMs themselves are likely imperfect.  It is well known that CLIP can exhibit erroneous agreements when using cosine similarity \citep{li2024erroneousagreementsclipimage}, and more generally that foundation models are subject to hallucinations \citep{huang2023survey} and shortcut learning \citep{yuan2024llms}.
Ultimately, \textit{it is unclear how VLM-provided supervision affects the learned concepts}.  This is precisely what our study aims to find.

\paragraph{The Concept Vocabulary}

Before proceeding, we briefly mention another important aspect of VLM-CBMs.
While most of these models can work with any prespecified vocabulary $\calT$, recent works suggest to generate it automatically by consulting a Large Language Model (LLM), such as ChatGPT \citep{hurst2024gpt}.  The LLM is simply asked to describe concepts that it deems useful for identifying specific classes \citep{oikarinen2023label,yang2023language,srivastava2024vlgcbm,feng2024bayesian}.
Even filtering out low-quality answers, there is still a chance that invalid concepts may crop up in the vocabulary, as shown in \citep{yang2023language} and in \ref{sec:concept_vocabulary_generation}.
Since we aim to assess the impact of VLM-supplied concepts, \textit{we use gold standard annotations as reference}.
Therefore, in our experiments we use the corresponding (gold standard) vocabulary for all VLM-CBMs, excluding \DNCBM{s} in the process as they do not support this mode of operation.

\section{Metrics for Concept Quality}
\label{sec:metrics}
We aim to assess to what extent VLM-CBM learn high-quality concepts.
Establishing formal conditions for concepts to be deemed interpretable is a difficult open problem \citep{marconato2023interpretability, zarlenga2023towards, barbiero2025neural} and beyond the scope of our paper.
We take a different route and ask whether the learned concepts satisfy a minimal set of reasonable desiderata, namely they are accurate (\Cref{sec:metric-acc}), avoid leakage (\Cref{sec:metric-leakage}), are disentangled (\Cref{sec:metric-disentanglement-compactness}) and avoid unwanted correlations (\Cref{sec:metric-impurity}).

\begin{figure}[!t]
    \centering
    \includegraphics[width=\textwidth]{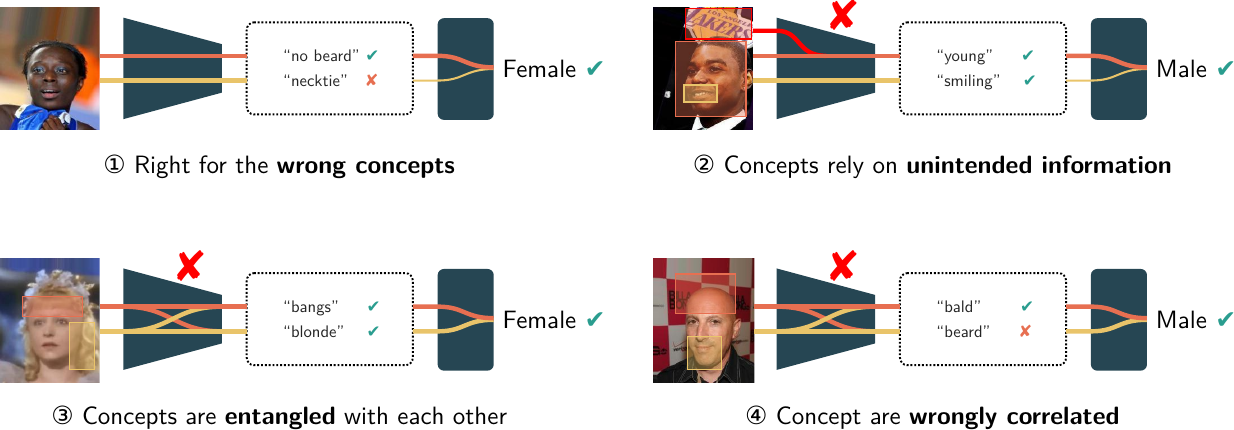}
    \caption{\textbf{Issues with CBM concepts}.  An illustration of the four issues with concept quality and usage affecting CBMs on the \CelebA task (cf. \cref{sec:experiments}):
    (1) Attaining high label accuracy does not prevent learning \textit{\textbf{inaccurate concepts}}.
    (2) CBMs can maximize label performance by learning \textbf{\textit{leaky concepts}} that include irrelevant contextual cues.
    (3) \textbf{\textit{Entangled concepts}} encode unwanted information about one another, affecting out-of-distribution behavor.
    (4) \textbf{\textit{Impure concepts}} encode unwanted correlations that do not exist among ground-truth concepts.
    }
    \label{fig:concept-issues}
\end{figure}

\subsection{Accurate Concept Predictions}
\label{sec:metric-acc}

The most natural requirement is that the learned concepts are \textbf{\textit{accurate}} \citep{koh2020concept}, \ie that they match their ground-truth values, irrespective of whether they were learned from expert or VLM supervision. 
Inaccurate concepts can still lead to correct decision-making, but they undermine the credibility of the CBMs' explanations.
For instance, if a VLM-CBM wrongly recognizes the presence of a \textit{``necktie''} and yet uses it to distinguish between male and female celebrities, cf. \cref{fig:concept-issues} (1), it leads to poor explanations of the decision-making.
This issue is well known, and indeed, poor concept predictions are the primary target for debugging, which can be done by manually labeling or intervening on concepts for misclassified instances \citep{koh2020concept, steinmann2024learning, zarlenga2024learning}.

While most works on VLM-CBMs measure accuracy \citep{oikarinen2023label, yang2023language, srivastava2024vlgcbm, rao2024discover}, this is unsuitable for unbalanced concepts (that, \eg occur rarely) and restricted to binary concepts.  However, as mentioned in \cref{sec:prelims}, different architectures use different concept supervision, from binary labels (\VLGCBM) to similarity scores (\LFCBM and \LABO).
Hence, we assess concept predictions by measuring the Area under the ROC curve, denoted \CAUC, averaged over all concepts in the bottleneck.  An additional benefit is that the AUC is robust to unbalanced concept classes, while the accuracy is not.

\subsection{Avoiding Concept Leakage}
\label{sec:metric-leakage}

High concept accuracy is not sufficient to ensure that the concepts possess human-aligned semantics.
Prior work has shown that (even accurate) concepts can suffer from \textit{\textbf{leakage}}\footnote{We are interested in leakage as it can be detrimental to interpretability \cite{mahinpei2021promises}, however it is still not clear if it should be avoided at all costs.  For instance, in some settings, leakage has been shown to improve the robustness of the model to interventions at the test time \cite{espinosa2023learning}.}, \ie they may unintentionally carry information that helps with label predictions but is semantically incongruent with the concept itself \citep{mahinpei2021promises, margeloiu2021concept, lockhart2022towards, havasi2022addressing, marconato2022glancenets}.
For instance, in \cref{fig:concept-issues} (2) the model maximizes the score of the correct (``\textit{male}'') class by encoding the name of a basketball team within the concept \textit{``young''}.

Unfortunately, leakage does not directly impact model performance, and as such it can be difficult to detect.
Prior works \citep{mahinpei2021promises, lockhart2022towards, marconato2022glancenets, havasi2022addressing} suggest to measure the retained amount of label accuracy when restricting the bottleneck to concepts that are known to be \textit{irrelevant} for the prediction task. 
In order to measure the overall leakage present in the whole bottleneck, we extend this idea as follows.
First, we sort all concepts by their \textit{ground-truth} Pearson correlation with the task label $y$, least correlated first.
Second, we train two linear SVMs \citep{cortes1995support} to predict the ground-truth label $y$ from the first $\ell$ concepts, one feeding on the ground-truth concept labels $\vc$ and the other on the predicted concepts $\hat \vc$, and measure the gap in their $F_1$-scores, \ie $\mathrm{gap}_\ell = F_1^{CBM}(Y; \ell) - F_1^{GT}(Y; \ell)$.
Intuitively, a linear classifier predicting the label from the first $\ell$ (\ie least correlated) concepts should perform very poorly if $\ell \ll k$; however, \textit{this is not the case if they leak information (about the label) encoded by remaining, most correlated ones}.
We repeat this step for $\ell = 1, \ldots, k$ and then record the average of all gaps:
\[
    \Leakage = \frac{1}{k \cdot Z} \sum_{\ell=1}^k \max( \mathrm{gap_\ell}, 0)
    \label{eq:leakage-gap}
\]
Here, 
$Z = 1 - \sum_{\ell=1}^k F_1^{GT}(Y; \ell)/k$ 
is a normalizing constant 
that ensures \Leakage has a maximum value equal to $1$.
\Leakage is zero if and only if all performance gaps are equal to or less than zero, \ie the learned concepts provide the same or less information about the label as the ground-truth ones.  Vice versa, a value close to 1 indicates that the learned concepts leak task-relevant information. 
The pseudo-code for its evaluation is confined in \cref{sec:leakage_calculation}.

\subsection{Disentanglement of Concepts}
\label{sec:metric-disentanglement-compactness}

Leakage occurs when concepts unintentionally encode spurious task-relevant information.
In some cases, this happens because they encode information about one another, as doing so can improve label accuracy.
Therefore, a key requirement is that learned concepts are \textbf{\textit{disentangled}} \citep{kazhdan2021disentanglement, marconato2023interpretability}, \ie manipulating the input to alter one concept should not affect the prediction of the others \citep{suter2019robustly}. 
E.g., in \cref{fig:concept-issues} (3) the presence of \textit{``bang''} is entangled with the \textit{``hair color''}.
Concepts that mix irrelevant factors are problematic as they do not match the semantics of their ground-truth counterpart, complicating interpretation \citep{marconato2023interpretability}.  For example, if the concept \textit{``bang''} contains information about the \textit{``hair color''}, it can behave unpredictably when this is switched on or off in the data \citep{montero2022lost}. 
Disentanglement is also a prerequisite for concept reuse, generalization, and supporting post-hoc alignment with users \citep{bortolotti2025shortcuts, fokkema2025sample}.

We measure disentanglement with \textsc{dci}, a popular metric that estimates the inter-relations between predicted and ground-truth concepts \citep{eastwood2018framework}.
It does so by training a non-linear model, \eg a random forest, predicting the ground-truth concepts from the learned ones and then computing relevance scores $R_{ij}$ that reflect how predictive the $i$-th learned concept is for the $j$-th ground-truth concept.
The \textsc{dci} disentanglement score $D_i$ for the concept $c_i$ is then defined as:
\begin{equation}
\textstyle
\label{eq:disentanglement}
    D_i := 1 - \big(-\sum_{j=1}^k P_{ij}\log_k P_{ij}\big)
\end{equation}
where $P_{ij} = R_{ij}/\sum_{\ell} R_{i\ell}$. The degree of disentanglement of the model is then given by computing a weighted average of all disentanglement scores, see \cref{sec:dci_framework}. 
\Disent has a maximum score of 1 and occurs when concepts are completely disentangled; instead, the score is 0 (worst case) if learned concepts equally depend on all ground-truth ones.

\subsection{{Unwanted} Correlations among Concepts} 
\label{sec:metric-impurity}

Disentanglement, as measured with \textsc{dci}, decreases with the degree to which learned concepts contain information about other concepts.

Nonetheless, in practice, learned concepts can be naturally correlated~\citep{zarlenga2023towards}, due to sampling bias of the data. 
For example, the concept \textit{``bald''} is likely to be correlated with the presence of \textit{``beard''}, so a model may capture these correlations.
However, if correlation among learned concepts exceeds the one in the data, the CBM may incur situations where \textit{``beard''} is predicted because of the absence of hair, as portrayed in \cref{fig:concept-issues} (4). 

The Oracle Impurity Score (OIS)~\citep{zarlenga2023towards} measures this ``impurity'' by penalizing whether unwanted correlations appear among the learned concepts. 
It does so by comparing the inter concept relations among ground-truth concepts with the inter concept relations among learned ones. 
Similar to \textsc{dci}, a non-linear model, \eg a random forest, is trained to predict the same ground-truth concepts from ground-truth concepts, obtaining a relevance matrix $R_{ij}^{(gt)}$. The same procedure is done by training a model from the same family to predict the ground-truth concepts from the model ones. This results to the same relevance matrix $R_{ij}$ accounted in \textsc{dci}. 

The divergence between these two matrices serves as an impurity metric, quantifying the amount of undesired information in the learned concepts:
\[
    \OIS = \frac{2}{k} \sum_{ij=1}^k |R_{ij} - R_{ij}^{(gt)}|^2
\]
An \OIS of $1$ indicates a complete misalignment between ground-truth and learned concepts, \ie the $i$-th concept representation can predict one or many other concepts, except the ones it supposed even when the concepts are independent, 
while an \OIS of $0$ denotes perfect alignment, meaning that the $i$-th learned concept does not capture any unwanted correlation.

\paragraph{Relationships between Metrics}

We point out that some of these metrics are partly correlated.
While concept accuracy is not a strong indicator of concept quality -- in the sense that even if it is perfect this does not rule out leakage \citep{mahinpei2021promises, margeloiu2021concept}, entanglement \citep{marconato2022glancenets, havasi2022addressing}, or impure correlations \citep{zarlenga2023towards} -- the amount of leakage does partly depend on whether concepts are entangled, and partly on whether they capture non-conceptual stylistic information \citep{marconato2023interpretability}.\footnote{Specifically, concepts that are both disentangled and independent from style entail no leakage, but the converse may not hold:  if a concept $c_i$ is \textit{non-linearly} entangled into another $c_j$ (according to \cref{eq:disentanglement}), a \textit{linear} layer may unable to extract useful information carried about the label by $c_i$ from $c_j$, thereby leading to zero leakage.}
In practice, especially in difficult classification tasks, the concept extractor may optimize for accuracy by exploiting wrongly correlated, entangled, or leaky concepts, precisely because doing so maximizes the amount of task-relevant information in the bottleneck \citep{zhang2024decoupling}.
Still, we empirically evaluate all four metrics so as to provide a fuller picture of VLM-CBMs concepts.

\section{Empirical Analysis}
\label{sec:experiments}

We tackle empirically the following research questions:
\begin{itemize}[leftmargin=2em]

    \item[\textbf{Q1}] Are expert and VLMs annotations comparable?
    
    \item[\textbf{Q2}] Are high label accuracy and high concept accuracy correlated in VLM-CBMs? 
    
    \item[\textbf{Q3}] Are VLM-CBM concepts of high-quality?

\end{itemize}

\paragraph{Data sets}
We consider three concept-annotated data sets for image classification.
\underline{\texttt{Sha}}{\tt p}\underline{\texttt{es3d}} \cite{kim2018disentangling} is a synthetic data set of renderings of $3$d objects with varying shape, color, orientation and background, for a total of $42$ different binary concepts. 
The ground-truth binary label $y \in \{0,1\}$ indicates whether a specific object (a \textit{red pill}) is present in the image.  In total, we use $48$k samples for training, $5$k for validation, and $30$k for testing.
\underline{\CelebA} \cite{liu2015faceattributes} contains about $200$k images (of size $178\times 218 \times 3$) of celebrity faces and provides manual annotations for $40$ binary concepts (hair color, presence of glasses, whether the person is smiling, \etc).  The model is asked to predict the ``gender'' of the celebrity from the remaining $39$ concepts.  We selected $25$k examples for training, $5$k for validation, and $20$k for testing. 
\underline{\CUB} \cite{wah2011caltech} is a bird classification task spanning $200$ different classes and $112$ concepts (bill shape, body color, feather pattern, \etc).
It consists of $4.8$k images for training, $1.2$k for validation, and $5.8$k for testing.
\paragraph{Architectures}
We evaluate three representative VLM-CBMs, namely \underline{\LABO} \citep{yang2023language}, \underline{\LFCBM} \citep{oikarinen2023label}, and \underline{\VLGCBM} \citep{srivastava2024vlgcbm}.  As baselines, we consider a standard \underline{\CBM} \citep{koh2020concept} and \underline{\method}, a CBM trained on concept labels gathered from a recent VLM balancing performance and efficiency.
Specifically, \method first queries LLAVA-Phi3 \citep{2023xtuner}, for each training image, about whether each concept in $\calT$ is present or not, and then fits a sparse linear layer on the resulting binary answers with a cross-entropy loss; see \cref{sec:naive_appendix} for further details.

For all models except \LABO, which has no backbone, we implement the image encoder using CLIP \citep{radford2021learning} (\texttt{ViT-B16} and \texttt{RN50} versions) or ResNet-18\citep{he2016deep} in both \SHAPES and \CelebA (the former for \LFCBM and \VLGCBM, the latter for \CBM and \method).
Following \citep{oikarinen2023label, srivastava2024vlgcbm}, in \CUB we use a ResNet-18 version pre-trained on CUB instead.  As we will show, this seemingly innocuous choice noticeably affects both label and concept quality.
In all cases, we implement the inference layer as a linear layer feeding on concept logits.
Following the original implementations, \LFCBM and \VLGCBM incentivize sparsity by using \texttt{GLM-SAGA} \citep{wong2021leveraging}, while \CBM and \method minimize the unregularized cross entropy loss and \LABO regularizes the linear layer by applying a softmax operator to the weight rows \citep{yang2023language}. 
All details regarding architectures and hyperparameter tuning are reported in \cref{sec:architectures,sec:hyperparameter_selection}.

\paragraph{Concept vocabularies}

All VLM-CBMs allow generating a task-specific concept vocabulary by querying an LLM (see \cref{sec:prelims}).  This procedure may yield invalid concepts \citep{srivastava2024vlgcbm} and more generally complicates comparing against gold-standard concept annotations.
To ensure a fair evaluation, in each dataset we fix the same vocabulary $\calT$ for all competitors.
The complete list of concept descriptions for each dataset can be found in \cref{sec:fixed_concept_set}.

\subsection{Q1: VLMs supervision does not match expert annotation}
\label{sec:q1}

We begin by assessing the quality of the VLM annotations.
To this end, we measure the macro precision and recall of the VLM annotations with respect to the gold standard annotations for all concepts in the vocabulary and report their average over the training examples in \cref{tab:results-VLMs}.
Overall, \textit{annotation quality differs substantially across VLMs and data sets}.
The best annotations are obtained on \CUB ($112$ concepts), with most VLMs achieving at least $53$\% precision and $58\%$ recall -- in fact, \texttt{G-DINO} attains high precision and recall above $90\%$ and \texttt{LLaVa} $92\%$ precision above and medium recall (about $66\%$).
VLM annotations in \CelebA ($39$ concepts) are noticeably worse -- the best option being \texttt{LLaVa}, with precision and recall above $67\%$ -- and very poor in \SHAPES (42 concepts):  \texttt{CLIP}, the best VLM, achieves precision and recall barely above $30\%$ and suffers from high variance.

We hypothesize that annotation quality depends on a combination of factors:
(\textit{i}) Whether the VLM has encountered concepts closely matching the queried textual description during training (\eg hair colors, eyeglasses).\footnote{We remark that understanding under what conditions enable VLMs to acquire high-quality concepts is still an open problem \citep{rajendran2024causal}.}
(\textit{ii}) Whether the input is in-distribution.  E.g., VLMs may struggle to annotate \SHAPES precisely because its synthetic images fall out-of-distribution.  It is in fact unlikely that VLMs would have trouble identifying concepts such as colors or shapes that likely appear in their training data.
While bigger VLM, such as Llama-vision 3.2 \citep{grattafiori2024llama}, might yield better annotations, they are not typically used for VLM-CBMs as they substantially increase computational costs.  We leave a more detailed study thereof to future work, while we provide few examples of automated annotations in \cref{fig:dino_annotations}. More details on the experiment can be found in \cref{sec:evaluating_vlm_annotations}.

\begin{table}[!t]
    \centering
    \caption{Agreement between ground-truth and VLMs annotations in terms of (macro) precision and recall, averaged across all concepts and examples.}
    \scriptsize
    \begin{tabular}{lllllllll}
\toprule
               & \multicolumn{2}{c}{\texttt{Shapes3d}} &  & \multicolumn{2}{c}{\texttt{CelebA}} &  & \multicolumn{2}{c}{\texttt{CUB200}} \\
               \cmidrule{2-3} \cmidrule{5-6} \cmidrule{8-9} 
               & \MacroPrecision $(\uparrow)$      & \MacroRecall $(\uparrow)$          &  & \MacroPrecision $(\uparrow)$          & \MacroRecall $(\uparrow)$         &  & \MacroPrecision $(\uparrow)$          & \MacroRecall $(\uparrow)$         \\
               \midrule

{\tt CLIP}           & $0.32\pm0.23$ & $0.30\pm 0.23$ & & $0.58 \pm 0.08$ & $0.65\pm 0.08$ & & $0.53\pm0.04$ & $0.58\pm 0.04$ \\ 
{\tt G-DINO}         & $0.18\pm0.07$ & $0.64\pm0.07$ & & $0.56\pm0.11$ & $0.54\pm0.11$ & & $0.97\pm 0.07$ & $0.90\pm 0.07$ \\ 
{\tt LLaVa}     & $0.13\pm0.06$ & $0.13\pm0.06$ & & $0.69\pm0.13$ & $0.67\pm0.13$ & & $0.92\pm0.08$ & $0.66\pm0.08$ \\ %
\bottomrule
\end{tabular}
    \label{tab:results-VLMs}
\end{table}

\subsection{Q2: VLM-CBMs output good labels but low-quality concepts}
\label{sec:q2}

Next, we evaluate VLM-CBMs predictions and concepts proper using $F_1$ score for the former (denoted \FY) and \CAUC for the latter (\cref{sec:metric-acc}).  The results for all models and data sets are reported in \cref{tab:all-results}.

A first key finding is that VLM-CBMs \textit{often achieve \FY comparable to the} \CBM \textit{baseline using substantially less accurate concepts}, as indicated by \CAUC.\footnote{The poor concept accuracy is likely due to poor VLM annotations, see \cref{sec:q1}.}
The gap is particularly noticeable in \SHAPES, where all VLM-CBMs have much lower \CAUC than \CBM, yet \LFCBM (and to a lesser extent \LABO) surprisingly achieve good $F_1$.  On the other hand, \method and \VLGCBM produce label predictions that look almost random, showing that the learned concepts carry very little to no information about the label.
The other two data sets paint a similar picture:
in \CelebA, \CAUC never crosses $50\%$ while \FY for all models is always above $95\%$, matching or even surpassing the \CBM baseline; and
in \CUB, the VLM-CBMs fare poorly at \CAUC -- the only exception being \VLGCBM, with $\CAUC = 79\%$ -- but all of them except \LABO achieve competitive \FY ($\ge 60\%$, close the \CBM's $69$\%).
Again, this suggests a disconnect between label and concept accuracy, and that \textit{maximizing the former provides no guarantees about the latter} \citep{bortolotti2025shortcuts}.

We readily acknowledge that for \LFCBM, \LABO, and \VLGCBM, our \FY results on \CUB are lower than those reported in the original publications \citep{oikarinen2023label, yang2023language, srivastava2024vlgcbm}. 
This is because the original evaluation used a broad set of LLM-generated concepts whose number can be up to ten thousand (a comparison of bottleneck sizes on \CUB dataset can be found in \cref{tab:original_bottleneck_size}). In contrast, \textit{we always employ the gold-standard concept vocabulary}.  Note that this is sufficient for inferring the correct labels, and indeed the associated manual concept annotations enable \CBM to perform well on all data sets in terms of \FY.  Our results indicate that \textit{restricting these models to the gold-standard vocabulary does
affect label accuracy compared to using larger vocabularies}. When contrasted with existing results, this leads us to hypothesize that a more extensive concept vocabulary allows the bottleneck to retain more information about label predictions  even if concepts do not correspond to anything meaningful \citep{srivastava2024vlgcbm} or they are randomly generated \citep{mahinpei2021promises}.

Finally, we attribute the lackluster performance of \LABO to the fact that it is the only model without a backbone pretrained on \CUB.  We believe such task-specific pre-training, which is common practice for this data set \citep{oikarinen2023label, srivastava2024vlgcbm}, brings a substantal performance advantage and as such it should be employed whenever feasible.

\begin{table}[!t] 
    \centering
    \caption{Results averaged over $5$ runs. Best results are in \textbf{bold} and second best are \underline{underlined}.}
    \scriptsize
    \begin{tabular}{llccccc}
        \toprule
        &
        {\sc Model}
            & \FY ($\uparrow$)
            & \CAUC ($\uparrow$)
            & \Leakage ($\downarrow$)
            & \Disent ($\uparrow$)
            & \OIS ($\downarrow$)
        \\
        \cmidrule{2-7}
        \multirow{5}{*}{
        \rotatebox{90}{\SHAPES}
        }
        & 
        \CBM
            & $\textbf{0.99} \pm 0.01$
            & $\textbf{0.99} \pm 0.01$
            & ${0.18} \pm 0.06$           
            & $\textbf{0.99} \pm 0.01$  
            & $\textbf{0.08} \pm 0.01$           
        \\
        &
        \method
            & $0.54  \pm 0.06$
            & $0.17  \pm 0.01$
            & $\underline{0.14} \pm 0.01$ 
            & $0.20 \pm 0.01$                   
            & $0.16 \pm 0.01$
        \\
        \cmidrule{2-7}
        &
        \LABO
            & $0.76 \pm 0.01$
            & $\underline{0.31} \pm 0.01$
            & $0.81 \pm 0.01$   
            & $0.28 \pm 0.01$   
            & $0.15 \pm 0.01$   
        \\
        &
        \LFCBM
            & $\underline{0.96} \pm 0.01$
            & $\underline{0.31} \pm 0.01$
            & ${0.64} \pm 0.01$   
            & $\underline{0.28} \pm 0.01$   
            & $\underline{0.15} \pm 0.01$   
        \\
        &
        \VLGCBM
            & $0.52 \pm 0.01$
            & $0.24 \pm 0.01$
            & $\textbf{0.06} \pm 0.10$   
            & $0.24 \pm 0.01$   
            & $0.16 \pm 0.01$
        \\
        \midrule
        \midrule
        \multirow{5}{*}{
        \rotatebox{90}{\CelebA}
        }
        &
        \CBM
            & $0.95 \pm 0.01$
            & $\textbf{0.75} \pm 0.01$
            & $\textbf{0.41} \pm 0.01$           
            & $\textbf{0.74} \pm 0.01$  
            & $\textbf{0.13} \pm 0.01$
        \\
        &
        \method
            & $0.95 \pm 0.01$
            & $\underline{0.51} \pm 0.01$
            & $0.99 \pm 0.01$   
            & $\underline{0.39} \pm 0.01$   
            & $\underline{0.16} \pm 0.01$
        \\
        \cmidrule{2-7}
        &
        \LABO
            & $0.97 \pm 0.01$
            & $0.38 \pm 0.01$
            & $\underline{0.42} \pm 0.01$   
            & $0.28 \pm 0.01$   
            & $0.18 \pm 0.01$   
        \\
        &
        \LFCBM
            & ${\textbf{0.99}} \pm 0.01$
            & $0.41 \pm 0.01$
            & $0.53 \pm 0.01$   
            & $0.29 \pm 0.01$   
            & $0.17 \pm 0.01$
        \\
        &
        \VLGCBM
            & $\underline{0.98} \pm 0.01$
            & $0.32 \pm 0.01$
            & $0.92 \pm 0.01$   
            & ${0.21} \pm 0.01$   
            & $0.20 \pm 0.01$
        \\
        \midrule
        \midrule
        \multirow{5}{*}{
        \rotatebox{90}{\CUB}
        }
        &
        \CBM
            & $\textbf{0.69} \pm 0.01$
            & $\textbf{0.91} \pm 0.01$
            & $0.10 \pm 0.02$           
            & $\textbf{0.75} \pm 0.01$  
            & $\textbf{0.09} \pm 0.01$  
        \\
        &
        \method
            & $0.66 \pm 0.01$
            & $0.43 \pm 0.03$
            & $\underline{0.02} \pm 0.01$   
            & $0.27 \pm 0.02$   
            & $0.15 \pm 0.01$
        \\
        \cmidrule{2-7}
        &
        \LABO
            & $0.27 \pm 0.01$
            & $0.25 \pm 0.01$
            & $\textbf{0.01} \pm 0.01$                   
            & $0.18 \pm 0.01$  
            & $0.15 \pm 0.01$                   
        \\
        &
        \LFCBM
            & $\underline{0.68} \pm 0.01$
            & $0.25 \pm 0.01$
            & ${\textbf{0.01}} \pm 0.01$                   
            & $0.19 \pm 0.01$  
            & $0.16 \pm 0.01$  
        \\
        &
        \VLGCBM
            & $0.60 \pm 0.01$
            & $\underline{0.79} \pm 0.01$
            & $\underline{0.02} \pm 0.01$               
            & $\underline{0.60} \pm 0.01$   
            & $\underline{0.10} \pm 0.01$   
        \\
        \bottomrule
    \end{tabular}
    \label{tab:all-results}
\end{table}

\begin{figure}[!t]
    \centering
    \begin{tabular}{cccc}
         \includegraphics[width=0.475\textwidth]{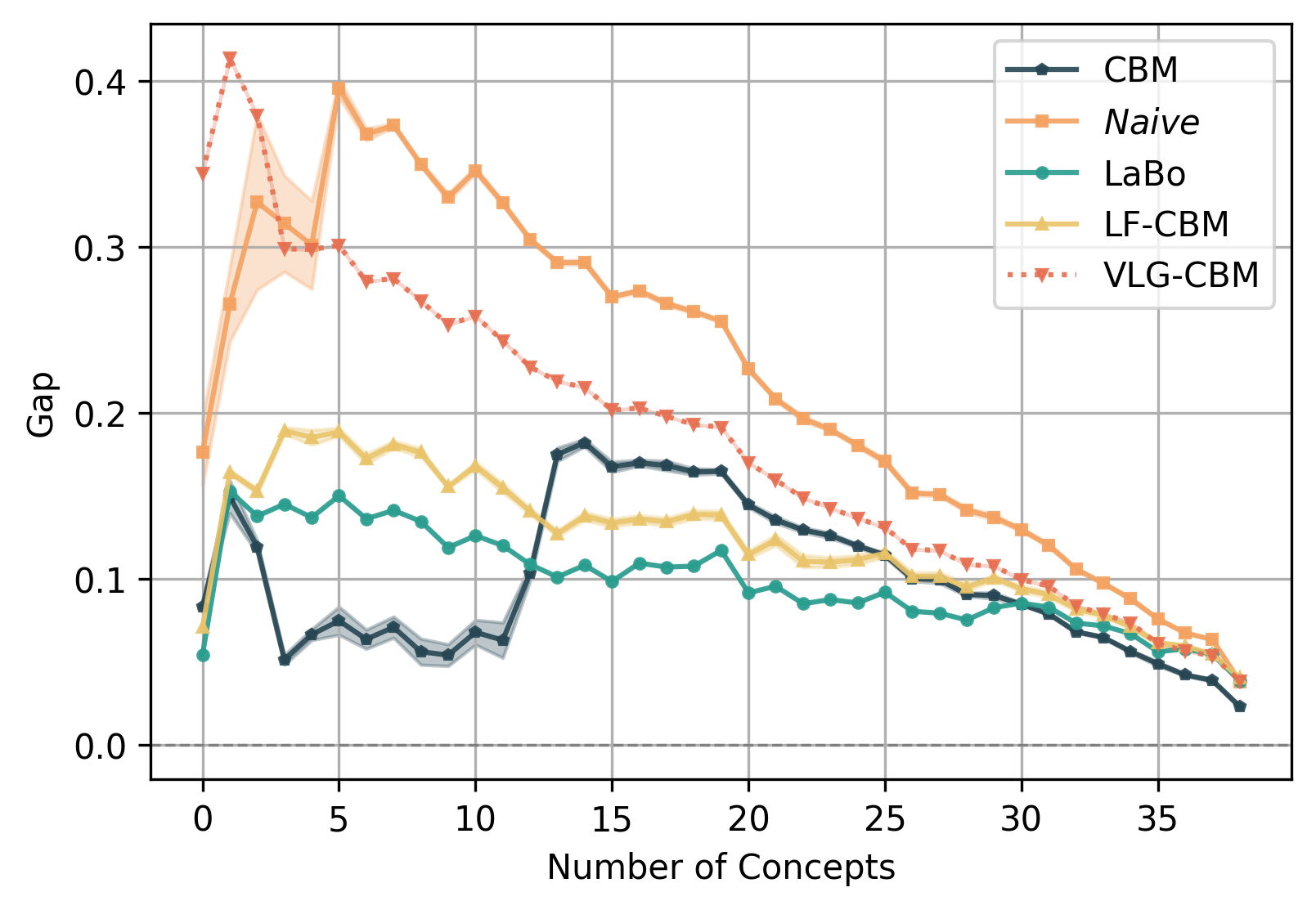}
         &
         \includegraphics[width=0.475\textwidth]{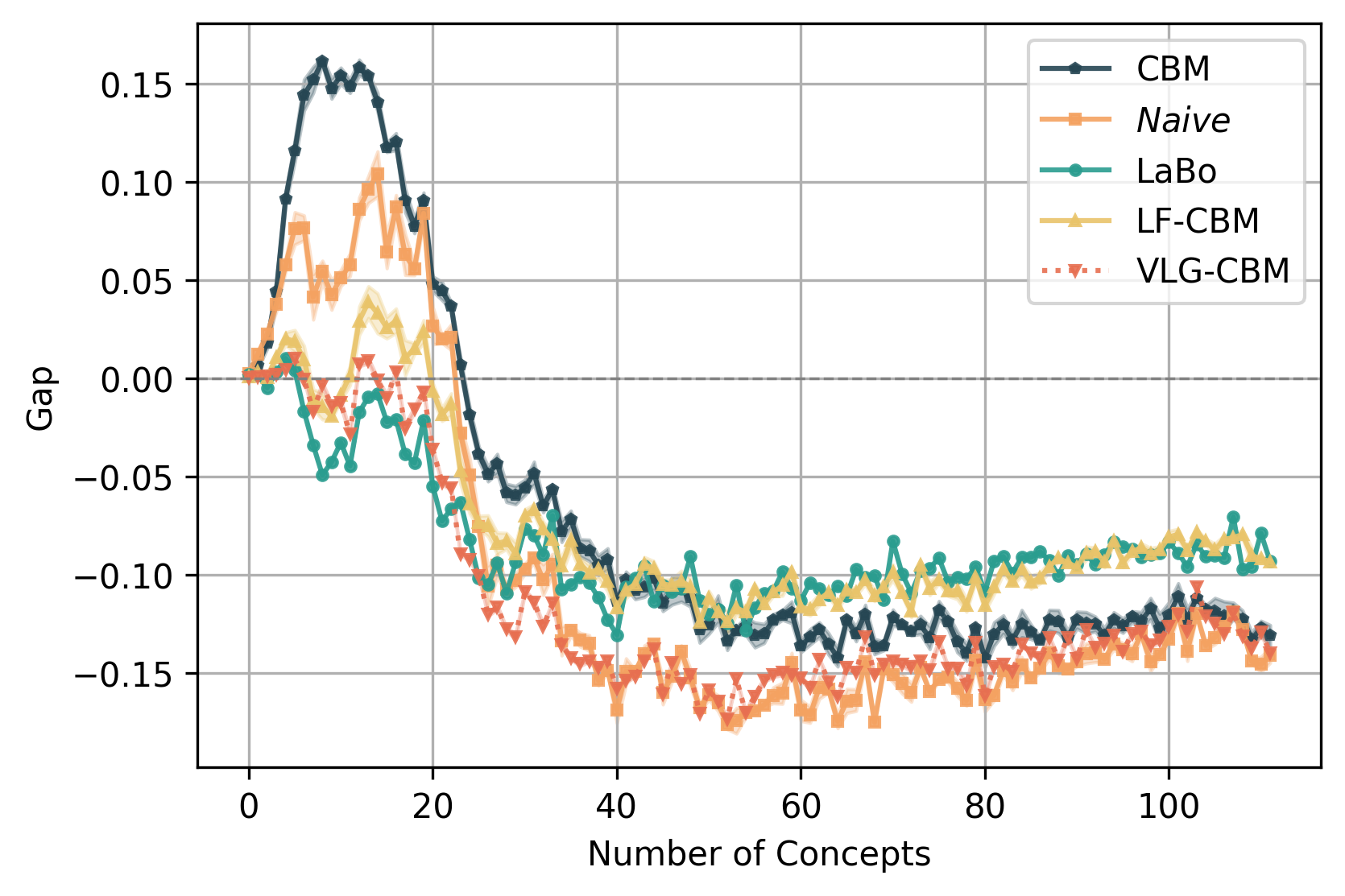}
    \end{tabular}
    \caption{\textbf{Gaps of the concept leakage test in \CelebA and \CUB}.
    The higher the gap, the more the subset of learned concepts leaks information for label prediction.
    (Left) Gaps on \CelebA upon varying the subset of concepts from $1$ to $39$. 
    All model concepts always display a non-negative gain in label $F_1$ compared to ground-truth concepts.
    (Right) Gaps on \CUB upon varying the subset of concepts from $1$ to $112$. Only \CBM is affected by leakage when considering smaller subsets of concepts.
    }
    \label{fig:leakage-records}
\end{figure}

\subsection{Q3: VLM-CBMs often leverage poor quality concepts}
\label{sec:q3}

We now turn to evaluating specific aspects of concept quality, beginning with leakage (\cref{sec:metric-leakage}).
Compatibly with existing findings \citep{mahinpei2021promises, margeloiu2021concept, havasi2022addressing, marconato2022glancenets}, the results in \cref{tab:all-results} indicate that \CBM \textit{is affected by leakage in all datasets}, with a maximum of $41\%$ in \CelebA and a minimum of $10\%$ in \CUB.
In \SHAPES, both CLIP-based models (\LFCBM and \LABO) are affected by leakage. \VLGCBM is the only method that sensibly avoids leakage in this dataset, despite faring poorly in label predictions. 
Note that a low $F_1(Y)$ does not imply leakage is absent, as demonstrated by the results of \method.\footnote{This can be noted from the plot on gaps in \cref{fig:leakage_gap_shapes3d}, where \method leaks a small amount of label information at early stages with a subset of a few irrelevant concepts. The gaps reduce at later stages, especially when adding all concepts, where, eventually, \method can no longer achieve label $F_1$ above the ground-truth.}

The trend is flipped in \CelebA: \LABO fares similarly to \CBM, \LFCBM drops by $11\%$, \method attains maximum leakage despite achieving an \CAUC higher than other VLM-CBMs, and \VLGCBM is highly affected by it, with $92\%$ leakage. 
In \cref{fig:leakage-records} (left), we report the gaps for \CelebA, which better spot on which subset of concepts (VLM) CBMs suffer leakage more. 
For example, the plots show that even if \CBM and \LABO have similar \Leakage ($41\%$ vs $42\%$), they display different behavior: \LABO displays a small amount of leakage in every concept, while \CBM reveals that only a few concepts are more leaky.
In particular, the concepts \textit{``big lips''} and \textit{``bald''} appear to introduce the most substantial amount of leakage on \CBM. 
Interestingly, we observe that gaps of all models are above zero when considering all concepts in the bottleneck.\footnote{This shows that \textit{extra, stylistic information} about labels is encoded in the bottleneck beyond the predefined ground-truth concepts. For example, as in \cref{fig:concept-issues} (2), the model could exploit textual information within the image (that is correlated to the label) to improve the prediction.}
The evaluation on \CUB portrays a different picture where only \CBM and \method are sensibly affected by leakage.  The high gaps for \CBM in \cref{fig:leakage-records} (right) show that a big proportion of label information is spread onto few concepts.  \VLGCBM and \LABO instead do not leak label information in the concepts, as visible for the always small or negative gaps.
 
Finally, we look at disentanglement and oracle impurity score (\cref{sec:metric-disentanglement-compactness,sec:metric-impurity}).
Overall, for all tested models, disentanglement is positively correlated with \CAUC while OIS is negative correlated with it.
In all datasets, \CBM attain the highest \Disent and the lowest \OIS thanks to expert annotation, whereas VLM-CBMs fare below the baseline \method in \SHAPES and \CelebA. On \CUB, \VLGCBM achieve higher \Disent and lower \OIS compared to other models, thanks to the higher quality of annotation of \texttt{G-DINO}. 
We analyze the DCI importance matrices $R_{ij}$ for  \CelebA on a single, optimal \FY run of each competitor, cf. \cref{fig:disent-vlm-cbms-celeba}. All VLM-CBMs drastically deviates from the behavior of \CBM, showing different entangled concepts based on the VLM they leverage. For example, \method displays high entanglement for the concept \textit{``goatee''} and \textit{``wearing lipstick''}, CLIP-based approaches show that \textit{``goatee''}, \textit{``no beard''}, and \textit{``smiling''} are the most entangled concepts, whereas \VLGCBM entangles several concepts, the most evident being \textit{``gray hair''}, and \textit{``pale skin''}.

\section{Related Work}
\label{sec:related-work}

\textbf{Concept-based models}  are a heterogeneous class of models that leverage learned concepts to glue together low-level perception with transparent high-level reasoning, see \citep{schwalbe2022concept, poeta2023concept, ji2025comprehensive} for recent overviews.
\textit{Unsupervised} models rely on concept discovery techniques, and foster interpretability via penalty terms architectural biases, such as \textit{sparsity} \citep{alvarez2018towards}, \textit{orthogonality} \citep{chen2020concept}, \textit{reconstruction} \citep{li2018deep} and \textit{prototypes} \citep{chen2019looks}.
Supervised CBMs differ in what information the encode in the concept bottleneck:  \textit{concept activations} \citep{koh2020concept}, \textit{embeddings} \citep{zarlenga2022concept}, \textit{probabilities} \citep{marconato2022glancenets, kim2023probabilistic, vandenhirtz2024stochastic}, or \textit{discrete activations} \citep{havasi2022addressing, lockhart2022towards}.
VLM-CBMs relax the need for expert supervision by leveraging pre-trained VLMs.

\vspace{1em}
\noindent
\textbf{The issue of interpretability in CBMs} is well-known.
Surprisingly, most works on VLM-CBMs do not focus on \textit{\textbf{concept quality}} \citep{oikarinen2023label, yang2023language, srivastava2024vlgcbm, rao2024discover, schrodi2024concept}. 
It is well-known that, however, the semantics of concepts learned from data can be misleading \citep{furby2023towards} and that even achieving high concept accuracy is not sufficient to ensure concepts are interpretable \citep{mahinpei2021promises, margeloiu2021concept, havasi2022addressing, marconato2022glancenets, barbiero2025neural}.
A number of metrics have been introduce to establish desirable properties of learned concepts. We focus on {\textit{leakage} \citep{havasi2022addressing, mahinpei2021promises}, \textit{disentanglement} \citep{marconato2022glancenets}, \textit{oracle impurity score}} \citep{zarlenga2023towards}, but other works have looked at \textit{completeness} \citep{yeh2020completeness, havasi2022addressing}, \textit{locality} \citep{raman2023concept}, or \textit{post-hoc alignment} \citep{mikriukov2023evaluating, bortolotti2025shortcuts}.
Techniques for improving concept quality build on causal insights \citep{bahadori2021debiasing} and on side-information \citep{sawada2022csenn, havasi2022addressing} obtained from either experts \citep{lertvittayakumjorn2020find, stammer2021right, bontempelli2023concept, teso2023leveraging} or foundation models \citep{srivastava2024vlgcbm}.
Provided that the extracted concepts are high-quality, the \textit{\textbf{classifier must also be high-quality}} to guarantee interpretable decision making.
Several works have indicated that  \textit{sparsity} is a central property \citep{alvarez2018towards, oikarinen2023label,barbiero2022entropy} and that can lead to least leakage of information \citep{srivastava2024vlgcbm}, while other works focus on making the classifier decision mechanism more \emph{understandable} by learning logic rules as in decision trees \citep{barbiero2024relational, barbiero2023interpretable}, or \emph{verifiable} \citep{debot2024interpretable}, or \textit{causally transparent} \citep{dominici2024causal, moreira2024diconstruct, de2025causally}. Since we focus specifically on concept quality, we leave to future work to examine how VLM-CBMs stimulate classifier quality.

\vspace{1em}
\noindent
\textbf{Beyond CBMs}.  The issue of concept quality extends beyond CBMs to other branches of AI, including Neuro-Symbolic AI and Foundation Models.
For instance, it was shown that the concepts learned by Neuro-Symbolic (NeSy) architectures can be compromised by shortcuts \citep{marconato2024not}, and these findings were recently extended to CBMs \citep{bortolotti2025shortcuts}.
A popular solution is, just like for CBMs, to leverage foundation models \citep{steinmann2024learning}.  Our work highlights that this strategy comes with its own set of pitfalls and cannot strictly guarantee the trustworthiness of the predictor.

\section{Conclusion}
\label{sec:conclusion}

Our results suggest that, despite significantly broadening the applicability of CBMs, VLMs are unfortunately not yet an accurate substitute for high-quality expert annotations.
Annotation quality translates not only to worse concept accuracy, but it also results in higher leakage, worse entanglement, and more impure concepts, depending on the dataset.
While it remains to be seen whether the quality of machine-generated concepts could match man-made ones, our results highlight that leakage remains an open problem even when using predefined (and sufficient) textual description of concepts.
In turn, this means that VLM-CBMs tend to be less interpretable than CBMs, at least in applications where a direct comparison with expert annotations is feasible.

With this work, we hope to help understand potential limitations of VLM-CBMs and to prompt researchers to place more emphasis on concept quality when developing future architectures.
We will conclude by pointing out possible ways forward.
One natural solution is, of course, to simply wait for VLMs to improve.  While this is not unlikely given recent trends, leveraging bigger models necessarily comes at a substantial computational cost.
We suggest that a more future-proof strategy, not necessarily orthogonal to the former, is to design VLM-CBMs that are more robust to low-quality concept annotations.
This could be achieved, for instance, by leveraging sound confidence estimates of the VLM's annotations during training \citep{abbasi2024believe}, exploiting well-known regularization terms and learning strategies which proved effective for regular CBMS \citep{bahadori2021debiasing, marconato2022glancenets, fokkema2025sample}, and actively seeking manual annotations for low-quality concepts \citep{zarlenga2024learning, steinmann2024learning}.

\section*{Acknowledgements}

Funded by the European Union. Views and opinions expressed are however those of the author(s) only and do not necessarily reflect those of the European Union or the European Health and Digital Executive Agency (HaDEA). Neither the European Union nor the granting authority can be held responsible for them. Grant Agreement no. 101120763 - TANGO.

\section*{Declarations}

The authors declare no competing interest.

\bibliography{main}

\begin{thebibliography}{10}

\bibitem{abbasi2024believe}
Yasin Abbasi~Yadkori et~al.
\newblock {To Believe or Not to Believe Your LLM: Iterative Prompting for
  Estimating Epistemic Uncertainty}.
\newblock {\em NeurIPS}, 2024.

\bibitem{alvarez2018towards}
David Alvarez~Melis and Tommi Jaakkola.
\newblock Towards robust interpretability with self-explaining neural networks.
\newblock {\em NeurIPS}, 2018.

\bibitem{ansel2024pytorch}
Jason Ansel et~al.
\newblock {PyTorch 2: Faster Machine Learning Through Dynamic Python Bytecode
  Transformation and Graph Compilation}.
\newblock In {\em ASPLOS}, 2024.

\bibitem{bahadori2021debiasing}
Mohammad~Taha Bahadori and David Heckerman.
\newblock Debiasing concept-based explanations with causal analysis.
\newblock In {\em ICLR}, 2021.

\bibitem{barbiero2022entropy}
Pietro Barbiero et~al.
\newblock Entropy-based logic explanations of neural networks.
\newblock In {\em AAAI}, 2022.

\bibitem{barbiero2023interpretable}
Pietro Barbiero et~al.
\newblock Interpretable neural-symbolic concept reasoning.
\newblock In {\em ICML}, 2023.

\bibitem{barbiero2024relational}
Pietro Barbiero et~al.
\newblock Relational concept bottleneck models.
\newblock {\em NeurIPS}, 2024.

\bibitem{barbiero2025neural}
Pietro Barbiero et~al.
\newblock Neural interpretable reasoning.
\newblock {\em arXiv:2502.11639}, 2025.

\bibitem{bontempelli2023concept}
Andrea Bontempelli et~al.
\newblock Concept-level debugging of part-prototype networks.
\newblock In {\em ICLR}, 2023.

\bibitem{bortolotti2025shortcuts}
Samuele Bortolotti et~al.
\newblock Shortcuts and identifiability in concept-based models from a
  neuro-symbolic lens.
\newblock {\em arXiv:2502.11245}, 2025.

\bibitem{calanzone2025logically}
Diego Calanzone et~al.
\newblock Logically consistent language models via neuro-symbolic integration.
\newblock In {\em ICLR}, 2025.

\bibitem{chauhan2023interactive}
Kushal Chauhan et~al.
\newblock Interactive concept bottleneck models.
\newblock In {\em AAAI}, 2023.

\bibitem{chen2019looks}
Chaofan Chen et~al.
\newblock This looks like that: Deep learning for interpretable image
  recognition.
\newblock {\em NeurIPS}, 2019.

\bibitem{chen2020concept}
Zhi Chen et~al.
\newblock Concept whitening for interpretable image recognition.
\newblock {\em Nature Machine Intelligence}, 2020.

\bibitem{2023xtuner}
XTuner Contributors.
\newblock Xtuner: A toolkit for efficiently fine-tuning llm.
\newblock \url{https://github.com/InternLM/xtuner}, 2023.

\bibitem{cortes1995support}
Corinna Cortes and Vladimir Vapnik.
\newblock Support-vector networks.
\newblock {\em Machine learning}, 1995.

\bibitem{de2025causally}
Giovanni De~Felice et~al.
\newblock Causally reliable concept bottleneck models.
\newblock {\em arXiv:2503.04363}, 2025.

\bibitem{debot2024interpretable}
David Debot et~al.
\newblock Interpretable concept-based memory reasoning.
\newblock {\em arXiv:2407.15527}, 2024.

\bibitem{dominici2024anycbmsturnblackbox}
Gabriele Dominici et~al.
\newblock Anycbms: How to turn any black box into a concept bottleneck model,
  2024.

\bibitem{dominici2024causal}
Gabriele Dominici et~al.
\newblock Causal concept graph models: Beyond causal opacity in deep learning.
\newblock {\em arXiv:2405.16507}, 2024.

\bibitem{dominici2024counterfactual}
Gabriele Dominici et~al.
\newblock Counterfactual concept bottleneck models.
\newblock {\em arXiv:2402.01408}, 2024.

\bibitem{eastwood2018framework}
Cian Eastwood and Christopher~KI Williams.
\newblock A framework for the quantitative evaluation of disentangled
  representations.
\newblock In {\em ICLR}, 2018.

\bibitem{espinosa2023learning}
Mateo Espinosa~Zarlenga et~al.
\newblock Learning to receive help: Intervention-aware concept embedding
  models.
\newblock {\em NeurIPS}, 2023.

\bibitem{zarlenga2024learning}
Mateo Espinosa~Zarlenga et~al.
\newblock Learning to receive help: Intervention-aware concept embedding
  models.
\newblock {\em NeurIPS}, 2024.

\bibitem{feng2024bayesian}
Jean Feng et~al.
\newblock Bayesian concept bottleneck models with llm priors.
\newblock {\em arXiv:2410.15555}, 2024.

\bibitem{fokkema2025sample}
Hidde Fokkema et~al.
\newblock Sample-efficient learning of concepts with theoretical guarantees:
  from data to concepts without interventions.
\newblock {\em arXiv:2502.06536}, 2025.

\bibitem{furby2023towards}
Jack Furby et~al.
\newblock Towards a deeper understanding of concept bottleneck models through
  end-to-end explanation.
\newblock In {\em Workshop on Representation Learning for Responsible
  Human-Centric AI @ AAAI}, 2023.

\bibitem{grattafiori2024llama}
Aaron Grattafiori et~al.
\newblock The llama 3 herd of models.
\newblock {\em arXiv preprint arXiv:2407.21783}, 2024.

\bibitem{havasi2022addressing}
Marton Havasi et~al.
\newblock Addressing leakage in concept bottleneck models.
\newblock In {\em NeurIPS}, 2022.

\bibitem{he2016deep}
Kaiming He et~al.
\newblock Deep residual learning for image recognition.
\newblock In {\em CVPR}, 2016.

\bibitem{higgins2018towards}
Irina Higgins et~al.
\newblock Towards a definition of disentangled representations.
\newblock {\em arXiv:1812.02230}, 2018.

\bibitem{huang2023survey}
Lei Huang et~al.
\newblock A survey on hallucination in large language models: Principles,
  taxonomy, challenges, and open questions.
\newblock {\em ACM TOIS}, 2023.

\bibitem{hurst2024gpt}
Aaron Hurst et~al.
\newblock Gpt-4o system card.
\newblock {\em arXiv:2410.21276}, 2024.

\bibitem{ismail2023concept}
Aya~Abdelsalam Ismail et~al.
\newblock Concept bottleneck generative models.
\newblock In {\em ICLR}, 2023.

\bibitem{ji2025comprehensive}
Yang Ji et~al.
\newblock A comprehensive survey on self-interpretable neural networks.
\newblock {\em arXiv:2501.15638}, 2025.

\bibitem{kazhdan2021disentanglement}
Dmitry Kazhdan et~al.
\newblock Is disentanglement all you need? comparing concept-based \&
  disentanglement approaches.
\newblock {\em arXiv:2104.06917}, 2021.

\bibitem{kim2018interpretability}
Been Kim et~al.
\newblock Interpretability beyond feature attribution: Quantitative testing
  with concept activation vectors.
\newblock In {\em ICML}, 2018.

\bibitem{kim2023probabilistic}
Eunji Kim et~al.
\newblock Probabilistic concept bottleneck models.
\newblock In {\em ICML}, 2023.

\bibitem{kim2018disentangling}
Hyunjik Kim and Andriy Mnih.
\newblock Disentangling by factorising.
\newblock In {\em ICML}, 2018.

\bibitem{kingma2014adam}
Diederik~P Kingma and Jimmy Ba.
\newblock Adam: A method for stochastic optimization.
\newblock {\em arXiv:1412.6980}, 2014.

\bibitem{koh2020concept}
Pang~Wei Koh et~al.
\newblock Concept bottleneck models.
\newblock In {\em ICML}, 2020.

\bibitem{laguna2024beyond}
Sonia Laguna et~al.
\newblock Beyond concept bottleneck models: How to make black boxes
  intervenable?
\newblock {\em NeurIPS}, 2024.

\bibitem{lai2024faithful}
Songning Lai et~al.
\newblock Faithful vision-language interpretation via concept bottleneck
  models.
\newblock In {\em ICLR}, 2024.

\bibitem{lertvittayakumjorn2020find}
Piyawat Lertvittayakumjorn et~al.
\newblock Find: human-in-the-loop debugging deep text classifiers.
\newblock In {\em EMNLP}, 2020.

\bibitem{li2018deep}
Oscar Li et~al.
\newblock Deep learning for case-based reasoning through prototypes: A neural
  network that explains its predictions.
\newblock In {\em AAAI}, 2018.

\bibitem{li2024erroneousagreementsclipimage}
Siting Li et~al.
\newblock On erroneous agreements of clip image embeddings.
\newblock {\em arXiv:2411.05195}, 2024.

\bibitem{liu2024grounding}
Shilong Liu et~al.
\newblock {Grounding DINO: Marrying DINO with Grounded Pre-Training for
  Open-Set Object Detection}.
\newblock In {\em ECCV}, 2024.

\bibitem{liu2015faceattributes}
Ziwei Liu et~al.
\newblock Deep learning face attributes in the wild.
\newblock In {\em ICCV}, 2015.

\bibitem{lockhart2022towards}
Joshua Lockhart et~al.
\newblock Towards learning to explain with concept bottleneck models:
  mitigating information leakage.
\newblock {\em arXiv:2211.03656}, 2022.

\bibitem{mahinpei2021promises}
Anita Mahinpei et~al.
\newblock Promises and pitfalls of black-box concept learning models.
\newblock In {\em Workshop on Theoretic Foundation, Criticism, and Application
  Trend of Explainable AI @ ICML}, 2021.

\bibitem{marconato2022glancenets}
Emanuele Marconato et~al.
\newblock {GlanceNets: Interpretabile, Leak-proof Concept-based Models}.
\newblock In {\em NeurIPS}, 2022.

\bibitem{marconato2023interpretability}
Emanuele Marconato et~al.
\newblock Interpretability is in the mind of the beholder: A causal framework
  for human-interpretable representation learning.
\newblock {\em Entropy}, 2023.

\bibitem{marconato2024not}
Emanuele Marconato et~al.
\newblock Not all neuro-symbolic concepts are created equal: Analysis and
  mitigation of reasoning shortcuts.
\newblock {\em NeurIPS}, 2024.

\bibitem{margeloiu2021concept}
Andrei Margeloiu et~al.
\newblock Do concept bottleneck models learn as intended?
\newblock {\em arXiv:2105.04289}, 2021.

\bibitem{mikriukov2023evaluating}
Georgii Mikriukov et~al.
\newblock {Evaluating the stability of semantic concept representations in CNNs
  for robust explainability}.
\newblock In {\em World Conference on Explainable Artificial Intelligence},
  2023.

\bibitem{montero2022lost}
Milton Montero et~al.
\newblock Lost in latent space: Examining failures of disentangled models at
  combinatorial generalisation.
\newblock {\em NeurIPS}, 2022.

\bibitem{moreira2024diconstruct}
Ricardo Moreira et~al.
\newblock Diconstruct: Causal concept-based explanations through black-box
  distillation.
\newblock {\em arXiv:2401.08534}, 2024.

\bibitem{oikarinen2023label}
Tuomas Oikarinen et~al.
\newblock Label-free concept bottleneck models.
\newblock In {\em ICLR}, 2023.

\bibitem{poeta2023concept}
Eleonora Poeta et~al.
\newblock Concept-based explainable artificial intelligence: A survey.
\newblock {\em arXiv:2312.12936}, 2023.

\bibitem{radford2021learning}
Alec Radford et~al.
\newblock Learning transferable visual models from natural language
  supervision.
\newblock In {\em ICML}, 2021.

\bibitem{rajendran2024causal}
Goutham Rajendran et~al.
\newblock From causal to concept-based representation learning.
\newblock {\em NeurIPS}, 2024.

\bibitem{raman2023concept}
Naveen Raman et~al.
\newblock Do concept bottleneck models obey locality?
\newblock In {\em XAI in Action: Past, Present, and Future Applications}, 2023.

\bibitem{rao2024discover}
Sukrut Rao et~al.
\newblock Discover-then-name: Task-agnostic concept bottlenecks via automated
  concept discovery, 2024.

\bibitem{sahu2022unpacking}
Pritish Sahu et~al.
\newblock Unpacking large language models with conceptual consistency.
\newblock {\em arXiv:2209.15093}, 2022.

\bibitem{sawada2022csenn}
Yoshihide Sawada and Keigo Nakamura.
\newblock C-senn: Contrastive self-explaining neural network.
\newblock {\em arXiv:2206.09575}, 2022.

\bibitem{sawada2022concept}
Yoshihide Sawada and Keigo Nakamura.
\newblock Concept bottleneck model with additional unsupervised concepts.
\newblock {\em IEEE Access}, 2022.

\bibitem{scholkopf2000new}
Bernhard Sch{\"o}lkopf et~al.
\newblock New support vector algorithms.
\newblock {\em Neural computation}, 2000.

\bibitem{scholkopf2021toward}
Bernhard Sch{\"o}lkopf et~al.
\newblock Toward causal representation learning.
\newblock {\em IEEE}, 2021.

\bibitem{schrodi2024concept}
Simon Schrodi et~al.
\newblock Concept bottleneck models without predefined concepts.
\newblock {\em arXiv:2407.03921}, 2024.

\bibitem{schwalbe2022concept}
Gesina Schwalbe.
\newblock Concept embedding analysis: A review.
\newblock {\em arXiv:2203.13909}, 2022.

\bibitem{shin2023closer}
Sungbin Shin et~al.
\newblock A closer look at the intervention procedure of concept bottleneck
  models.
\newblock In {\em ICML}, 2023.

\bibitem{srivastava2024vlgcbm}
Divyansh Srivastava et~al.
\newblock {VLG-CBM: Training Concept Bottleneck Models with Vision-Language
  Guidance}.
\newblock In {\em NeurIPS}, 2024.

\bibitem{stammer2021right}
Wolfgang Stammer et~al.
\newblock {Right for the Right Concept: Revising Neuro-Symbolic Concepts by
  Interacting with their Explanations}.
\newblock In {\em CVPR}, 2021.

\bibitem{steinmann2024learning}
David Steinmann et~al.
\newblock Learning to intervene on concept bottlenecks.
\newblock In {\em ICML}, 2024.

\bibitem{suter2019robustly}
Raphael Suter et~al.
\newblock Robustly disentangled causal mechanisms: Validating deep
  representations for interventional robustness.
\newblock In {\em ICML}, 2019.

\bibitem{teso2023leveraging}
Stefano Teso et~al.
\newblock Leveraging explanations in interactive machine learning: An overview.
\newblock {\em Frontiers in Artificial Intelligence}, 2023.

\bibitem{vandenhirtz2024stochastic}
Moritz Vandenhirtz et~al.
\newblock Stochastic concept bottleneck models.
\newblock {\em arXiv:2406.19272}, 2024.

\bibitem{wah2011caltech}
Catherine Wah et~al.
\newblock The caltech-ucsd birds-200-2011 dataset.
\newblock 2011.

\bibitem{wong2021leveraging}
Eric Wong et~al.
\newblock Leveraging sparse linear layers for debuggable deep networks.
\newblock In {\em ICML}, 2021.

\bibitem{yang2023language}
Yue Yang et~al.
\newblock Language in a bottle: Language model guided concept bottlenecks for
  interpretable image classification.
\newblock In {\em CVPR}, 2023.

\bibitem{yeh2020completeness}
Chih-Kuan Yeh et~al.
\newblock On completeness-aware concept-based explanations in deep neural
  networks.
\newblock {\em NeurIPS}, 2020.

\bibitem{yuan2024llms}
Yu~Yuan et~al.
\newblock {Do LLMs overcome shortcut learning? An evaluation of shortcut
  challenges in large language models}.
\newblock {\em arXiv:2410.13343}, 2024.

\bibitem{yuksekgonul2023post}
Mert Yuksekgonul et~al.
\newblock Post-hoc concept bottleneck models.
\newblock In {\em ICLR}, 2023.

\bibitem{zarlenga2022concept}
Mateo~Espinosa Zarlenga et~al.
\newblock Concept embedding models: Beyond the accuracy-explainability
  trade-off.
\newblock In {\em NeurIPS}. 2022.

\bibitem{zarlenga2023towards}
Mateo~Espinosa Zarlenga et~al.
\newblock Towards robust metrics for concept representation evaluation.
\newblock In {\em AAAI}, 2023.

\bibitem{zhang2024decoupling}
Rui Zhang et~al.
\newblock The decoupling concept bottleneck model.
\newblock {\em PAMI}, 2024.

\end{thebibliography}
\bibliographystyle{plain}

\newpage
\begin{appendices}

\newpage

\section{Experimental Details}
All experiments were implemented using Python 3 and Pytorch \citep{ansel2024pytorch}. The code is available at \href{https://github.com/debryu/CQA}{https://github.com/debryu/CQA}.

\subsection{Architectures}
\label{sec:architectures}
\newcommand{\bottleneck}{\texttt{\#Conc.}\xspace}
\newcommand{\classes}{\texttt{\#Class}\xspace}
\newcommand{\crossentropy}{\texttt{XENT}\xspace}
\newcommand{\svm}{\texttt{SVM}\xspace}

We list here the description of all architectures used in our experiments, where, for ease of reading, we omit the details about the backbones we used. 
For each component we specify its name, input, output, what it produces (i.e. the output of a layer could simply be an intermediate process, and if so the field is empty, or return important information such as logits or probabilities) and lastly if the layer/component is frozen or being trained. 
We also specify how a particular component is trained, where the options are:
\begin{enumerate}
    \item \crossentropy: standard PyTorch cross entropy loss
    computed between ground-truth and predicted values. 
    \item \svm: we use Support Vector Machine \cite{scholkopf2000new} (specifically the scikit learn \href{https://scikit-learn.org/stable/modules/generated/sklearn.svm.LinearSVC.html}{implementation}) to fit a linear classifier. 
    \item \SAGA: we minimize an elastic net loss using the \texttt{GLM-SAGA} solver created by \cite{wong2021leveraging}, in order to obtain a sparse linear classifier.
\end{enumerate}

\begin{table}[!h]
    \centering
    \caption{\CBM architecture.}
    \label{tab:cbm_architecture}
    \begin{tabular}{lccccl}
        \toprule
        Input & Type & Output & Produces  & Trained & with\\
        \midrule
        $(224,224,3)$ & ResNet18 & (1000) &  & \cmark & 
        \\
        $(1000)$ & Linear & (\bottleneck) &  Concept Logits  & \cmark & \crossentropy\\
        (\bottleneck) & Sigmoid & (\bottleneck) &  Concept Probs  & \xxmark &\\
        (\bottleneck) & Linear & (\classes) &  Class Logits  & \cmark & \SAGA\\
        \bottomrule
    \end{tabular}
\end{table}

\begin{table}[!h]
    \centering
    \caption{\LABO architecture.}
    \label{tab:labo_architecture}
    \begin{tabular}{lccccl}
        \toprule
        Input & Type & Output & Produces  & Trained & with\\
        \midrule
        $(224,224,3)$ & CLIP ViT-B/16 & (768) &  & \xxmark & \\
        $(768)$,$(768,\bottleneck)$\footnotemark & Dot Product & (\bottleneck) &    & \xxmark & \\
        (\bottleneck) & Normalization & (\bottleneck) & Concept Logits  & \xxmark & \\
        (\bottleneck) & Linear & (\classes) &  Class Logits  & \cmark & \crossentropy\\
        \bottomrule
    \end{tabular}
\end{table}

\footnotetext{$(768)$ is the CLIP image embedding, while $(768,\bottleneck)$ are the textual concepts encoded using CLIP.}

\begin{table}[!h]
    \centering
    \caption{\method architecture.}
    \label{tab:method_architecture}
    \begin{tabular}{lccccl}
        \toprule
        Input & Type & Output & Produces  & Trained & with\\
        \midrule
        $(224,224,3)$ & ResNet18 & (1000) &  & \cmark & 
        \\
        $(1000)$ & Linear & (\bottleneck) &  Concept Logits  & \cmark & \crossentropy\\
        (\bottleneck) & Sigmoid & (\bottleneck) &  Concept Probs  & \xxmark &\\
        (\bottleneck) & Linear & (\classes) &  Class Logits  & \cmark & \svm\\
        \bottomrule
    \end{tabular}
\end{table}

\begin{table}[!h]
    \centering
    \caption{\LFCBM architecture.}
    \label{tab:lfcbm_architecture}
    \begin{tabular}{lccccl}
        \toprule
        Input & Type & Output & Produces  & Trained & with\\
        \midrule
        $(224,224,3)$ & CLIP ViT-B/16 & (768) &  & \xxmark & \\
        $(768)$ & Linear & (\bottleneck) &  Concept Logits  & \cmark & \crossentropy\\
        (\bottleneck) & Linear & (\classes) &  Class Logits  & \cmark & \SAGA\\
        \bottomrule
    \end{tabular}
\end{table}

\begin{table}[!h]
    \centering
    \caption{\VLGCBM architecture.}
    \label{tab:vlgcbm_architecture}
    \begin{tabular}{lccccl}
        \toprule
        Input & Type & Output & Produces  & Trained & with\\
        \midrule
        $(224,224,3)$ & CLIP ViT-B/16 & (768) &  & \xxmark & \\
        $(768)$ & Linear & (\bottleneck) &    & \cmark & \crossentropy\\
        (\bottleneck) & Normalization & (\bottleneck) & Concept Logits  & \xxmark & \\
        (\bottleneck) & Linear & (\classes) &  Class Logits  & \cmark & \SAGA\\
        \bottomrule
    \end{tabular}
\end{table}

For \CUB, the ResNet18 output shape is $200$, matching the number of classes, and so the following linear layer will have input of dimension 200 while keeping unvaried the other components of the architecture.

Pre-trained models are dowloaded from \href{https://pytorch.org/vision/main/models.html}{PyTorch} and \href{https://github.com/osmr/imgclsmob}{imgclsmob}.

\subsection{\method}
\label{sec:naive_appendix}

We consider a fixed concept dictionary $\calT$ consisting of a total of $k$ concept textual descriptions.
According to the description of VLM-CBMs training \cref{sec:architectures},
our proposed baselines consist of the following three steps:
\begin{enumerate}
    \item Gathering concept annotation for all input-output pairs from the training set by querying a \texttt{LLaVa} model to provide a binary label if any of the concepts in $\calT$ is present. 
    
    \item We use the collected binary concept annotation to train the concept extractor $f$.

    \item By freezing the concept extractor, a linear layer $g$ is trained to predict the classes from concept activations. 
\end{enumerate}
To execute the first step, similarly to \citep{srivastava2024vlgcbm}, we prompt the VLM with the following script:
\begin{verbatim}
    "This is an image of {class_name}. 
    Does the image contain {prefix}{obj}{suffix}?
    Please reply only with 'Yes' or 'No'."
\end{verbatim}
where $\verb|class_name|$ is the class associated to the input sample and $\verb|obj|$ is the concept being queried. 
To adapt to different datasets, we added $\verb|prefix|$ and $\verb|suffix|$, which can be empty strings or customized to enrich the prompt. For example, in \CelebA we use the prefix \textit{"a person with"}, so that the full prompt ends up being "\verb|Does the image contain {a person with} {Moustache}?|". 

When the model does not follow the prompt with its answer, we append after the response
\begin{verbatim}
    "Given this information, my 'Yes' or 'No' answer is:"
\end{verbatim}
If the response from the model is still not satisfactory, we set the concept annotation value to 0.
The VLM is run locally, using \href{https://ollama.com}{ollama}.

\subsection{Leakage Calculation}
\label{sec:leakage_calculation}

To establish each concept's relevance to label prediction, we compute the Pearson correlation coefficient between each ground-truth concept and the label. 
Then, both ground-truth and predicted concepts are sorted by increasing correlation with the label, so that highly correlated concepts are placed last. 
Next, we use the iterative algorithm in \cref{algo:leakage} to implement the procedure detailed in \cref{sec:metric-leakage}. 
\begin{algorithm}
\caption{Compute F1 Gap Between Predicted and Ground-Truth Concepts}
\begin{algorithmic}[1]
\State gaps = $[~]$
\For{$i = 1$ to $k$}
    \State $\hat{C}_{train} \gets \text{predicted concepts on train set}$
    \State $\hat{C}_{test} \gets \text{predicted concepts on test set}$
    \State SVM.train($\hat{C}_{train}[:,1:i], y_{\text{train}}$)
    \State predictions $\gets$ SVM.predict($\hat{C}_{test}[1:i]$)
    \State F1\_predicted $\gets$ compute\_F1(predictions, $y_{\text{test}}$)
    \State $C_{train} \gets \text{ground truth concepts on train set}$
    \State $C_{test} \gets \text{ground truth concepts on test set}$
    \State SVM.train($C_{train}[:,1:i], y_{\text{train}}$)
    \State predictions $\gets$ SVM.predict($C_{test}[1:i]$)
    \State F1\_ground\_truth $\gets$ compute\_F1(predictions, $y_{\text{test}}$)
    \State gaps.append($F_1$\_predicted $-$ $F_1$\_ground\_truth)
\EndFor
\State \textbf{return:} gaps
\Comment{List of $F_1$ gaps for each step}
\end{algorithmic}
\label{algo:leakage}
\end{algorithm}

\subsection{Hyperparameter selection}
\label{sec:hyperparameter_selection}

During training, we apply early stopping when a model's validation loss does not decrease after 16 consecutive epochs.

We conducted a grid search over different sets of hyperparameters and determined the best combination by monitoring the label $F_1$ score on the validation set with wandb.
As for the optimizer, we used Adam \citep{kingma2014adam}.
Besides more common hyperparameters, 
we use a flag \texttt{unfreeze} to control the number of layers to unfreeze from the ResNet backbone (starting from the last one). 
The flag \texttt{balanced} is used to handle dataset imbalances when training the concept bottleneck.

\par \textbf{\CBM model parameters} are set as: 
\texttt{-epochs=20}, 
\texttt{-unfreeze=5}, 
\texttt{-lr=0.001}, 
\texttt{-batch\_size=512}, 
\texttt{-backbone=resnet18}, 
\texttt{-optimizer=adamw}, 
\texttt{-balanced}.
\par \textbf{\method model parameters} are set as: 
\texttt{-epochs=20}, 
\texttt{-unfreeze=5}, 
\texttt{-lr=0.001}, 
\texttt{-predictor=svm},
\texttt{-c\_svm=1},
\texttt{-batch\_size=512}, 
\texttt{-backbone=resnet18}, 
\texttt{-optimizer=adamw}, 
\texttt{-balanced},
\texttt{-ollama\_model=llava-phi3}.
\par \textbf{\LABO model parameters} are set as: 
\texttt{-lr=0.01}, \texttt{-batch\_size=258}, 
\texttt{-backbone=clip\_RN50}, 
\texttt{-feature\_layer=layer4}, 
\texttt{-clip\_name=ViT-B/16}.
\par \textbf{\LFCBM  model parameters} are set as: \texttt{-lr=0.01}, \texttt{-batch\_size=258}, \texttt{-backbone=clip\_RN50}, \texttt{-feature\_layer=layer4}, \texttt{-clip\_name=ViT-B/16}, \texttt{-proj\_steps=1000}, \texttt{clip\_cutoff=0.0}, \texttt{-interpretability\_cutoff=0.0}.\footnote{The last two hyperparameters are set to zero to prevent unwanted concept filtering.}

\par \textbf{\VLGCBM model parameters} are set as: 
\texttt{-backbone=clip\_RN50},
\texttt{-feature\_layer=layer4},
\texttt{-crop\_to\_concept\_prob=0.01},
\texttt{-cbl\_confidence\_threshold=0.15},
\texttt{-cbl\_hidden\_layers=3},
\texttt{-cbl\_epochs=8},
\texttt{-cbl\_lr=0.001},
\texttt{-val\_split=0.1}.

\par For the models that require \textbf{\SAGA} \citep{wong2021leveraging} we use the following parameters:
\texttt{-glm\_alpha=0.99}, \texttt{-glm\_step\_size=0.1}, \texttt{-n\_iters=2000}, \texttt{-lam=0.0007}, \texttt{-saga\_batch\_size=256}.

Lastly, when running experiments on \CUB we use
\texttt{-backbone=resnet18\_cub} for \CBM, \method, \LFCBM and \VLGCBM. For \LFCBM and \VLGCBM we also specify
\texttt{-feature\_layer=features.final\_pool} which is the last layer of the pre-trained ResNet backbone.
All the other hyperparameters are kept the same, except for \textbf{\VLGCBM parameters} which are: 
\texttt{-backbone=resnet18\_cub},
\texttt{-feature\_layer=features.final\_pool},
\texttt{-crop\_to\_concept\_prob=0.1},
\texttt{-cbl\_confidence\_threshold=0.15},
\texttt{-cbl\_hidden\_layers=0},
\texttt{-cbl\_epochs=35},
\texttt{-cbl\_lr=0.0005}.
\texttt{-lam=0.0002}.
\newpage

\subsection{DCI framework}
\label{sec:dci_framework}

The disentanglement metric is defined as the weighted average over all disentanglement scores $D_i$ appearing in \cref{eq:disentanglement}, where $i \in \{1,\ldots, k\}$:
\begin{equation}
    D = \sum_{i=1}^{n} \rho_iD_i, \quad \text{where} \;
    \rho_i = \frac{\sum_{i=1}^N R_{ij}}{\sum_{\ell=1}^N\sum_{j=1}^K R_{\ell j}}
\end{equation}

\subsection{Evaluating VLM annotations}
\label{sec:evaluating_vlm_annotations}
To evaluate VLM annotations we use macro average precision and recall measured on predicted concepts versus ground-truth ones. While \texttt{G-DINO} and \method produce boolean annotations (i.e. a concept is either 0 or 1), this is not true for \texttt{CLIP} embeddings. To overcome this, we train a logistic regressor model for each concept (using \texttt{CLIP} embeddings as input), and then assign 0 or 1 to the concept based on the output of that model.

We also take into consideration whether some concepts are mutually exclusive, such as in \SHAPES where concept 1 through 10 are the one-hot encoding of the color, so only one will be active. With that in mind, when evaluating \texttt{CLIP} annotations on \SHAPES we evaluate concept activation separately for each group of concepts (wall color, background color, object color, size and shape), setting as \textit{active} only the concept with highest activation withing the group.

\subsection{Concept Vocabulary}
\label{sec:concept-vocabularies}

\subsubsection{Concept set generation}
\label{sec:concept_vocabulary_generation}

When the concept vocabulary is not available,
a common choice is to query a Large Language Model to create a set of textual descriptions of the concepts $\calT$. However, one known limitation is that an LLM can produce invalid concepts \citep{yang2023language}. 
In general, a high-quality concept vocabulary generation is difficult because the concept names need to avoid being overly specific or narrow for the task.
In \cref{tab:invalid_concepts}, we report some examples of invalid concepts in the generated vocabularies for \CUB. These concept descriptions are taken from 
the original evaluation of the methods we tested.
\begin{table}[h]
    \centering

    \caption{Examples of invalid concepts from the \CUB vocabulary generated using LLMs. The concepts are taken from the filtered set of the original implementation \citep{oikarinen2023label,yang2023language,srivastava2024vlgcbm}.}

    \renewcommand{\arraystretch}{1.5} 
    \begin{tabular}{|c|p{5cm}|p{5cm}|} 
        \hline
        \textbf{Method} & \multicolumn{2}{c|}{\textbf{Generated Concepts}} \\
        \hline
        & \textbf{Irrelevant} & \textbf{Non-visual} \\
        \hline
        \textbf{\LFCBM} \citep{oikarinen2023label} & \textsf{"a Mohs hardness of 10"} & {\textsf{"loud, rasping voice"}} \\
                    &  & {\textsf{"a loud, harsh cry"}} \\
                    &  & {\textsf{"a loud, high-pitched song"}} \\
        \hline
        \textbf{\LABO} \citep{yang2023language} & \textsf{"hollow trees or stumps"} & {\textsf{"series of high, piping notes"}} \\
                    & \textsf{"adds color and life to our forests"} & {\textsf{"song is a trill"}} \\
                    & \textsf{"named after anna mary robertson, better known as grandma moses"} & {\textsf{"cawing can be quite loud and annoying"}} \\
        \hline
        \textbf{\VLGCBM} \citep{srivastava2024vlgcbm} & \textsf{"a trackpad or mouse"} & {\textsf{"a loud, melodious song"}} \\
                    & & {\textsf{"loud, rasping voice"}} \\
                    & & {\textsf{"a loud caw"}} \\
        \hline
    \end{tabular}
    \label{tab:invalid_concepts}
\end{table}

\subsubsection{Fixed Concept Set}
\label{sec:fixed_concept_set}
Our objective is to use a vocabulary of concepts that can be shared across different models, which also includes access to ground-truth values. With that in mind, we took the textual description of the ground-truth concepts, with some minor preprocessing such as removing underscores for better understanding.
We list below the complete sets of concepts used in our experiments for the \CUB, \CelebA, and \SHAPES datasets.

\clearpage
\newpage
\noindent
\textbf{\large \CUB concept set}
\label{set:cub_concept_set}

\begin{minipage}[t]{0.225\textwidth} 
    \footnotesize
    dagger beak         \\
    hooked seabird beak  \\
    all-purpose beak \\
    cone beak \\
    brown wing \\
    grey wing \\
    yellow wing \\
    black wing \\
    white wing \\
    buff wing \\
    brown upperparts \\
    grey upperparts \\
    yellow upperparts \\
    black upperparts \\
    white upperparts \\
    buff upperparts \\
    brown underparts \\
    grey underparts \\
    yellow underparts \\
    black underparts \\
    white underparts \\
    buff underparts \\
    solid breast \\
    striped breast \\
    multi-colored breast \\
    brown back \\
    grey back \\
    yellow back \\
    black back \\     
\end{minipage}
\hfill
\begin{minipage}[t]{0.225\textwidth} 
    \footnotesize
    white back \\
    buff back \\
    notched tail \\
    brown upper-tail \\
    grey upper-tail \\
    black upper-tail \\
    white upper-tail \\
    buff upper-tail \\
    eyebrow head \\
    plain head \\
    brown breast     \\
    grey breast     \\
    yellow breast     \\
    black breast     \\
    white breast     \\
    buff breast     \\
    grey throat     \\
    yellow throat     \\
    black throat     \\
    white throat     \\
    buff throat     \\
    black eye     \\
    beak length about-- \\
    -- the same as head     \\
    beak length shorter--\\ 
    -- than head     \\   
    blue forehead     \\
    brown forehead     \\
    grey forehead     \\
\end{minipage}   
\hfill
\begin{minipage}[t]{0.225\textwidth} 
    \footnotesize
    yellow forehead     \\
    black forehead     \\
    white forehead     \\
    brown under-tail     \\
    grey under-tail     \\
    black under-tail     \\
    white under-tail     \\
    buff under-tail     \\
    brown nape     \\
    grey nape     \\
    yellow nape     \\
    black nape     \\
    white nape     \\
    buff nape     \\
    brown belly     \\
    grey belly     \\
    yellow belly     \\
    black belly     \\
    white belly     \\
    buff belly  \\
    rounded wings      \\
    pointed wings     \\
    size small     \\
    size medium     \\
    very small size     \\
    duck-like     \\
    perching-like     \\
    solid back     \\
    striped back     \\
\end{minipage}
\hfill
\begin{minipage}[t]{0.225\textwidth} 
    \footnotesize
    multi-colored back     \\
    solid tail     \\
    striped tail     \\
    multi-colored tail     \\
    solid belly     \\
    brown primary color     \\
    grey primary color     \\
    yellow primary color     \\
    black primary color     \\
    white primary color     \\
    buff primary color     \\
    grey leg     \\
    black leg     \\
    buff leg     \\
    grey beak     \\
    black beak     \\
    buff beak     \\
    blue crown     \\
    brown crown     \\
    grey crown     \\
    yellow crown     \\
    black crown     \\
    white crown     \\
    solid wing     \\
    spotted wing     \\
    striped wing     \\
    multi-colored wing \\
\end{minipage}


\noindent
\textbf{\large \SHAPES concept set}
\label{set:shapes3d_concept_set}

\begin{minipage}[t]{0.3\textwidth} 
    \footnotesize
    red floor       \\
    orange floor    \\
    yellow floor    \\
    green floor     \\
    lime floor      \\
    azure floor    \\
    blue floor    \\
    navy blue floor    \\
    purple floor    \\
    pink floor    \\
    red wall    \\
    orange wall    \\
    yellow wall    \\
    green wall 
\end{minipage}
\hfill
\begin{minipage}[t]{0.3\textwidth} 
    \footnotesize
   lime wall   \\
    azure wall   \\
    blue wall   \\
    navy blue wall   \\
    purple wall   \\
    pink wall   \\
    red object   \\
    orange object   \\
    yellow object   \\
    green object   \\
    lime object   \\
    azure object   \\
    blue object   \\
    navy blue object   
\end{minipage}   
\hfill
\begin{minipage}[t]{0.3\textwidth} 
    \footnotesize
    purple object   \\
    pink object   \\
    miniscule object   \\
    very small object   \\
    small object   \\
    medium size object   \\
    slightly big object   \\
    big object   \\
    very big object   \\
    enormous object   \\
    cube shaped object   \\
    cylinder shaped object   \\
    sphere shaped object   \\
    pill shaped object   
\end{minipage}

\vspace{1em}

\noindent
\textbf{\large \CelebA concept set}
\label{set:celeba_concept_set}
\vspace{0.5em}

\begin{minipage}[t]{0.3\textwidth} 
    \footnotesize
    5 o Clock Shadow    \\
    Arched Eyebrows    \\
    Attractive    \\
    Bags Under Eyes    \\
    Bald    \\
    Bangs    \\
    Big Lips    \\
    Big Nose    \\
    Black Hair    \\
    Blond Hair    \\
    Blurry    \\
    Brown Hair    \\
    Bushy Eyebrows    
\end{minipage}
\hfill
\begin{minipage}[t]{0.3\textwidth} 
    \footnotesize
    Chubby      \\
    Double Chin    \\
    Eyeglasses    \\
    Goatee    \\
    Gray Hair    \\
    Heavy Makeup    \\
    High Cheekbone    \\
    Mouth Slightly Open    \\
    Mustache    \\
    Narrow Eyes    \\
    No Beard    \\
    Oval Face    \\
    Pale Skin
\end{minipage}   
\hfill
\begin{minipage}[t]{0.3\textwidth} 
    \footnotesize
    Pointy Nose          \\
    Receding Hairline      \\
    Rosy Cheeks      \\
    Sideburns      \\
    Smiling      \\
    Straight Hair      \\
    Wavy Hair      \\
    Wearing Earrings      \\
    Wearing Hat      \\
    Wearing Lipstick      \\
    Wearing Necklace      \\
    Wearing Necktie      \\
    Young
\end{minipage}
\clearpage

\subsection{Concept Sets in other works}
\label{sec:original_concepts}
As a reference, we provide the number of concepts used in the original implementation of the methods we use in our experiments.

\begin{table}[!h]
    \centering
    \caption{Number of selected concepts for \CUB dataset.}
    \label{tab:original_bottleneck_size}
    \begin{tabular}{lc}
        \toprule
            Model & N. Concepts \CUB \\
        \midrule
            \CBM \cite{koh2020concept}   & 112 \\
            \LFCBM \cite{oikarinen2023label}& 370 \\
            \LABO \cite{yang2023language}& $\ge 2000$\\
            \VLGCBM \cite{srivastava2024vlgcbm} &   723 \\
        \bottomrule
    \end{tabular}
\end{table}

\section{Additional experimental results}
\label{sec:additional-exp-supp}

\begin{figure}[!h]
    \centering
    \includegraphics[width=0.9\textwidth]{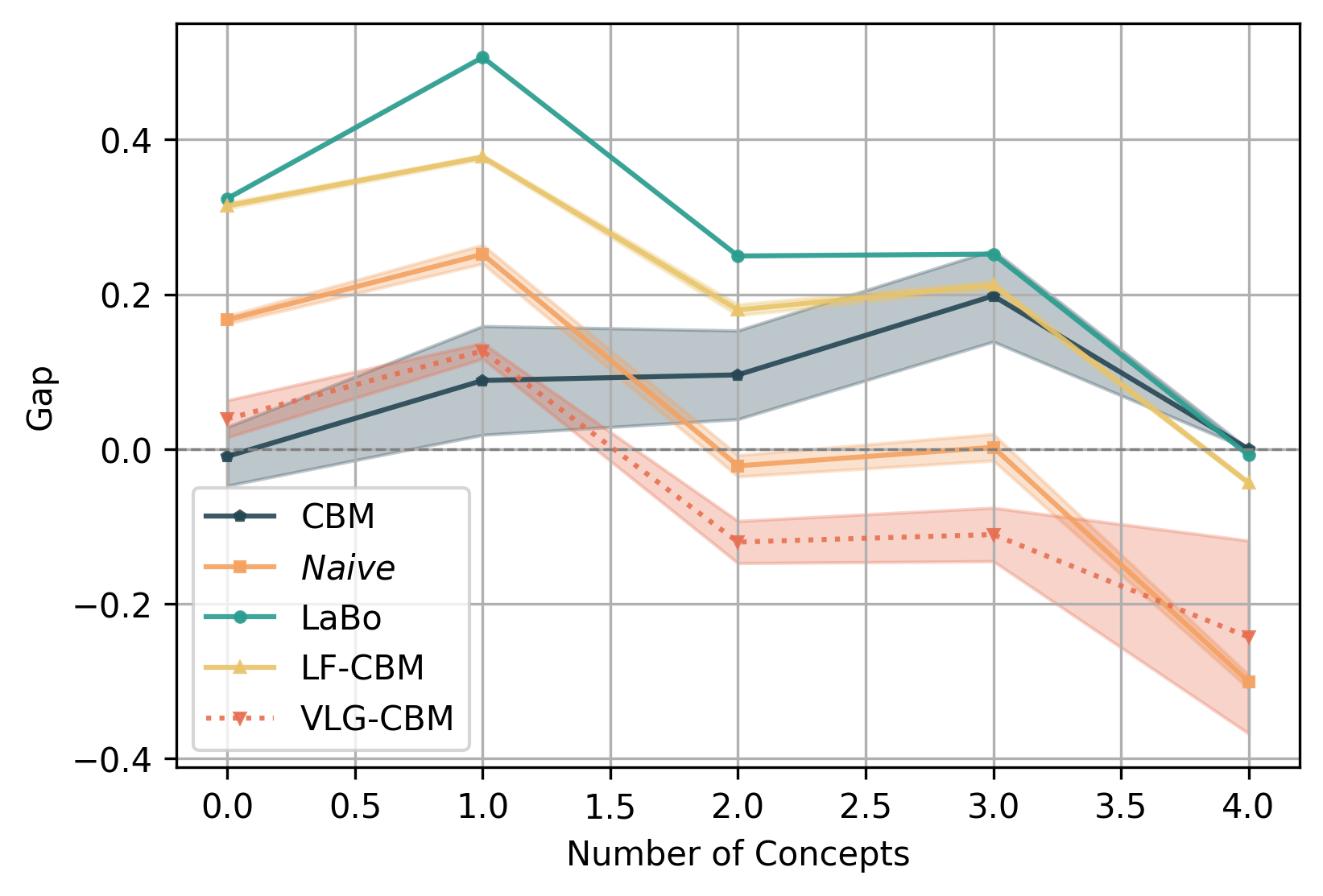}
    \caption{Gaps in concept leakage test on \SHAPES dataset. Negative values indicate that the classifier using predicted concepts has lower $F_1$ score than the same which has access to ground-truth concept annotations.}
    \label{fig:leakage_gap_shapes3d}
\end{figure}

\begin{figure}[!t]
    \vspace{0.5em}
    \centering
    
    \includegraphics[width=0.97\textwidth]{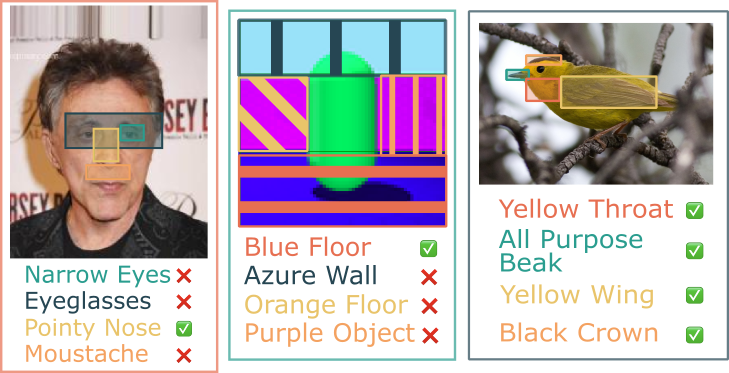}
    \vspace{7pt}
     \caption{Annotations obtained using \texttt{G-DINO} on the 3 different datasets. On \CelebA (left) and \SHAPES (center) the VLM makes several mistakes highlighted by red crosses. On the other hand, the model does a good job annotating \CUB (right).}
    \label{fig:dino_annotations}
\end{figure}

\subsection{Disentanglement}
\begin{figure}[!t]
    \centering
    \begin{tabular}{ccr}
        \CBM & \method  \\
         \includegraphics[width=0.3\textwidth]{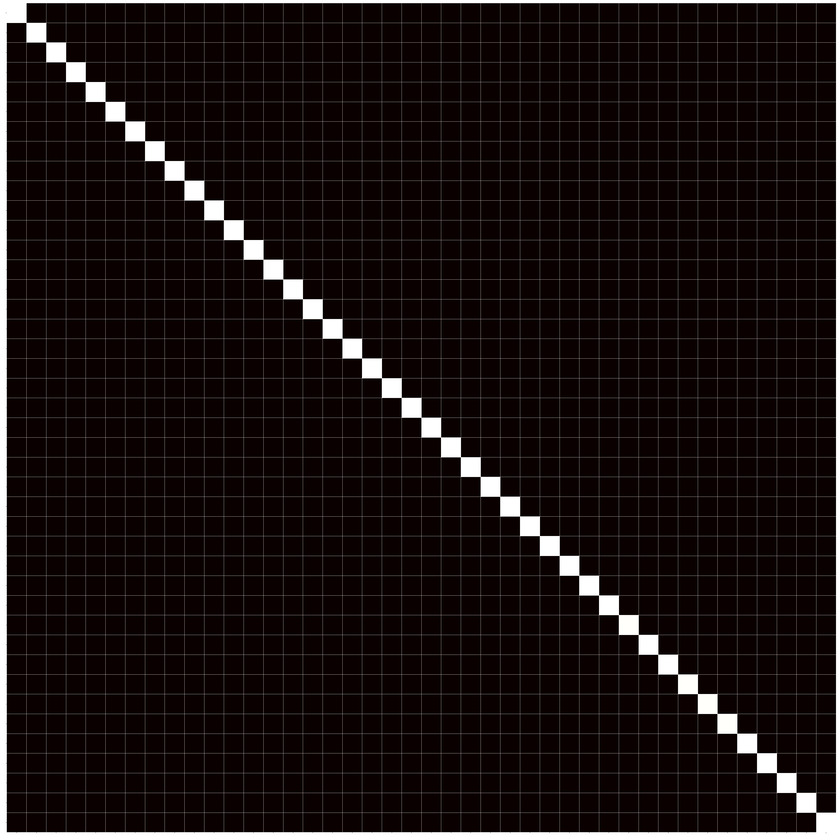}
         & 
         \includegraphics[width=0.3\textwidth]{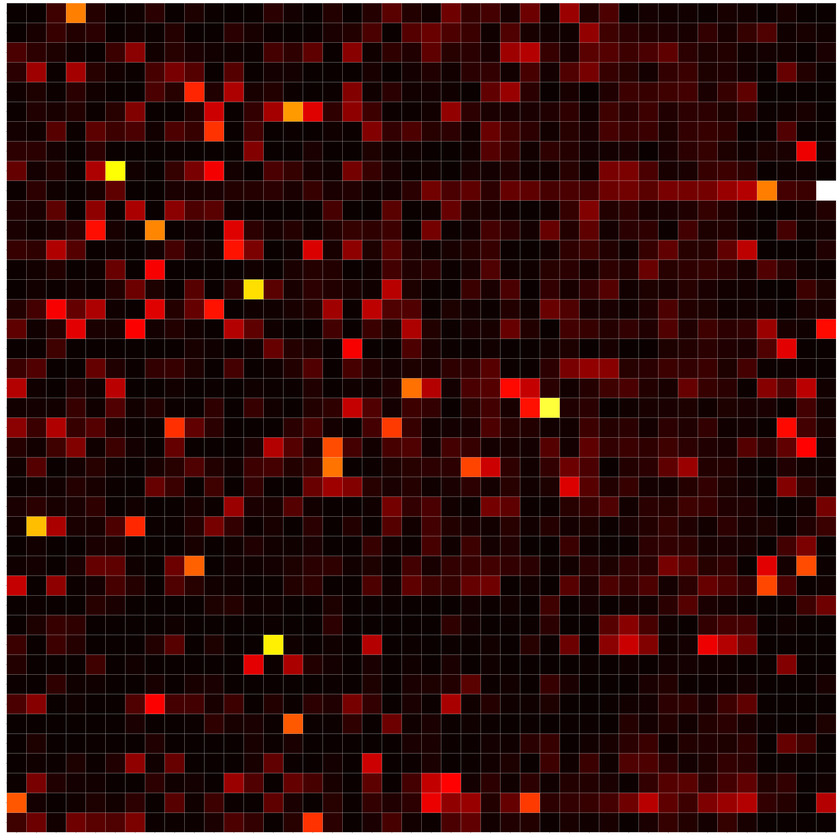}
         & 
         \includegraphics[height=5cm]{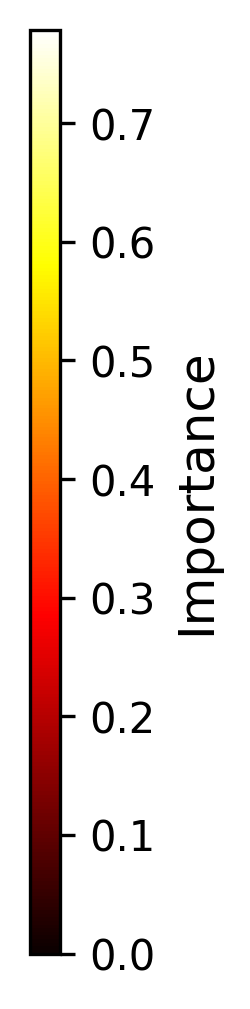}
    \end{tabular}

    \begin{tabular}{ccc}
        \LFCBM & \LABO &  \VLGCBM \\
         \includegraphics[width=0.3\textwidth]{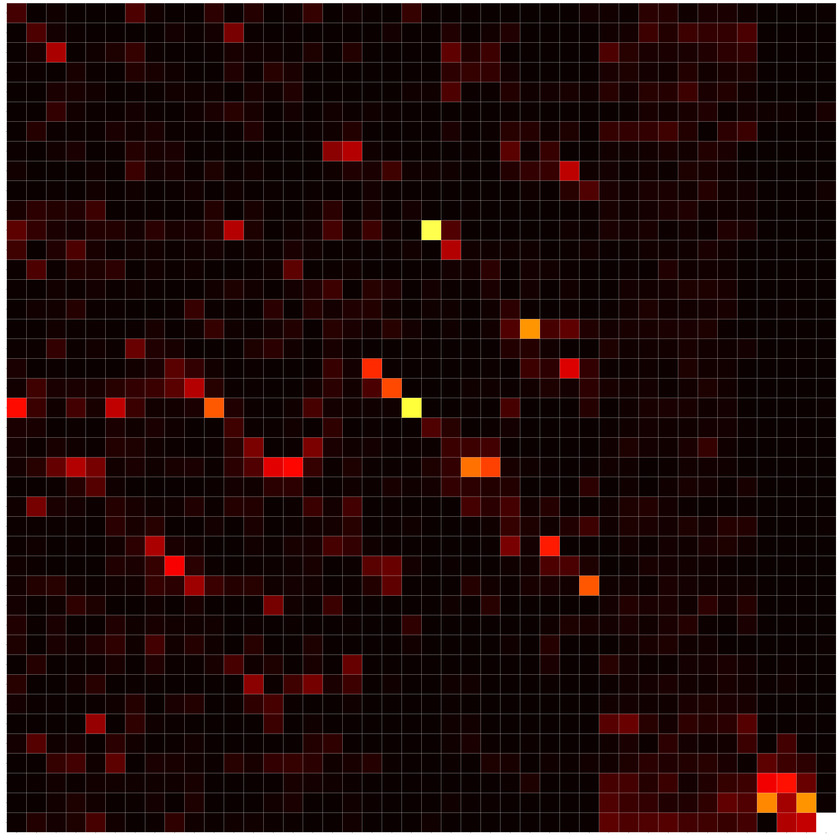}
         & 
         \includegraphics[width=0.3\textwidth]{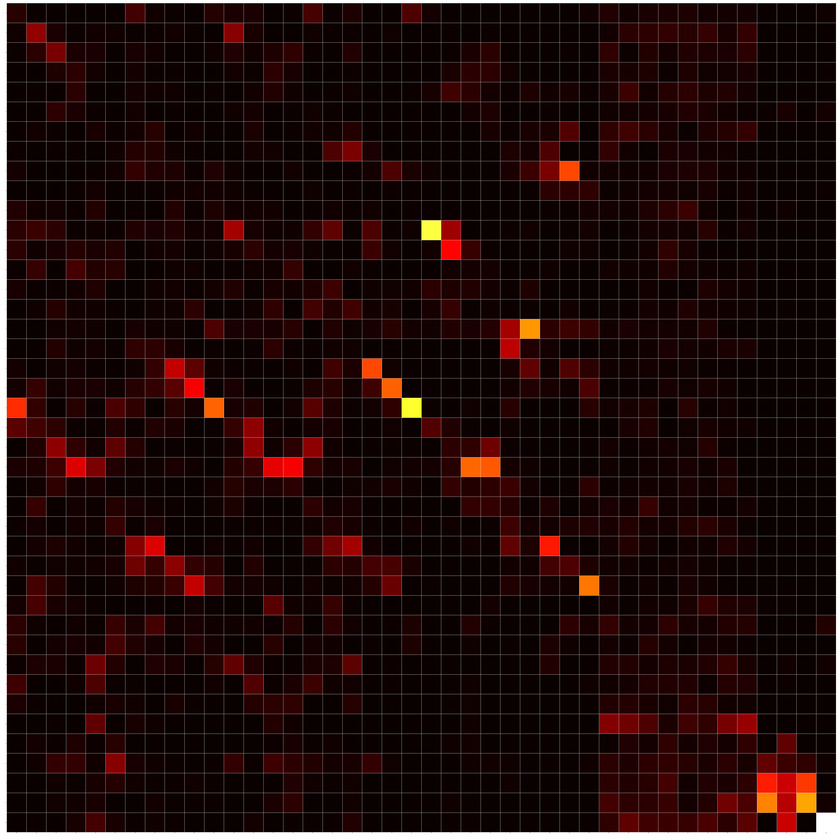}
         & 
         \includegraphics[width=0.3\textwidth]{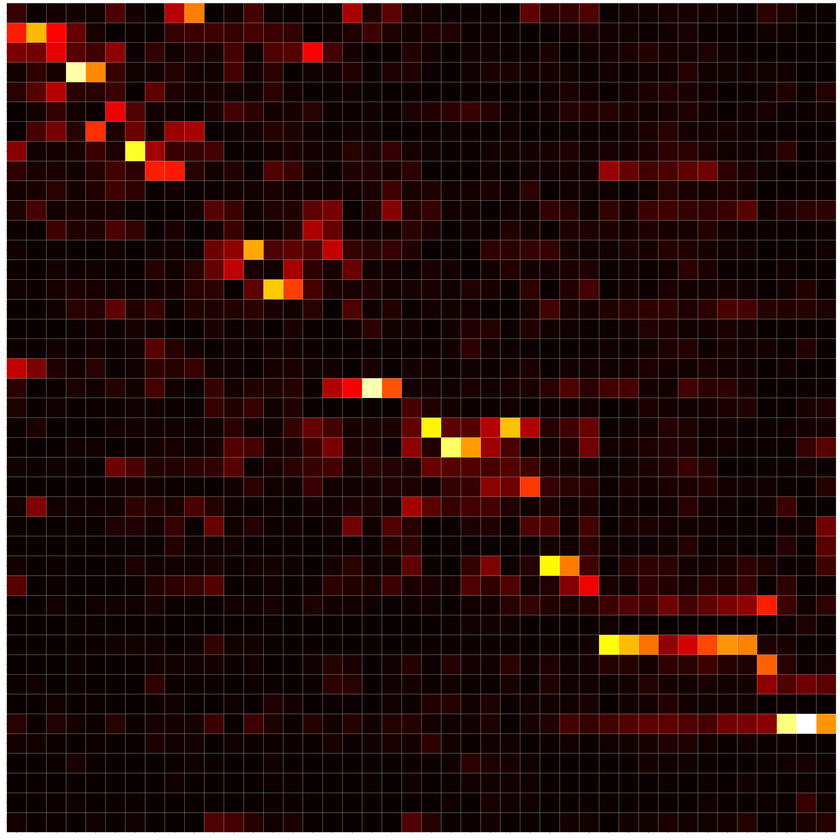}
    \end{tabular}
    \caption{\textbf{Importance matrices in DCI on \SHAPES}. High disentanglement is achieved when, for each row, only one square is active. For reference, a perfect DCI matrix is a diagonal matrix.}
    \label{fig:disent-vlm-cbms-shapes3d}
\end{figure}

\begin{figure}[!t]
    \centering
    
    \begin{tabular}{ccr}
        \CBM & \method  \\
         \includegraphics[width=0.3\textwidth]{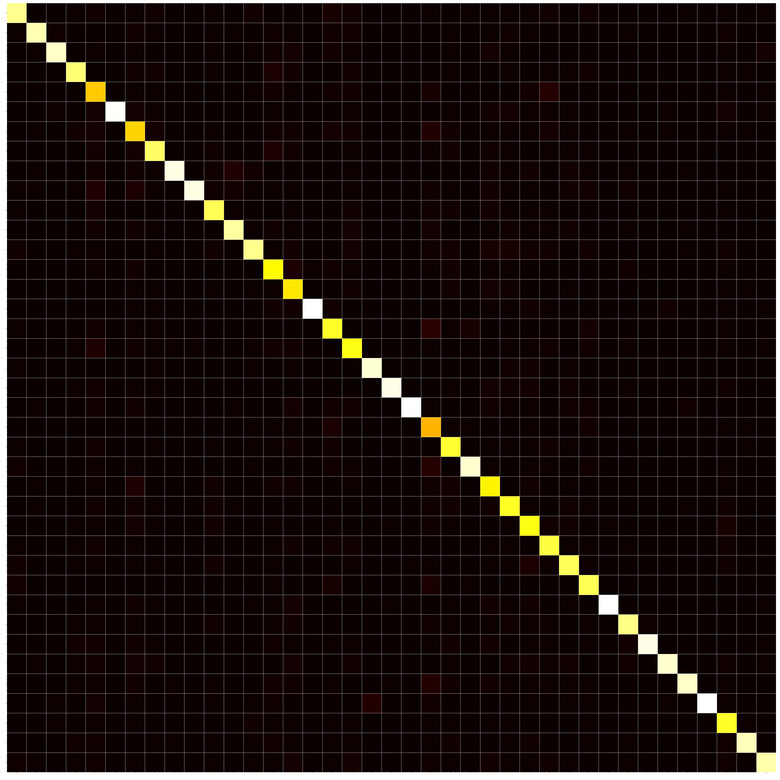}
         & 
         \includegraphics[width=0.3\textwidth]{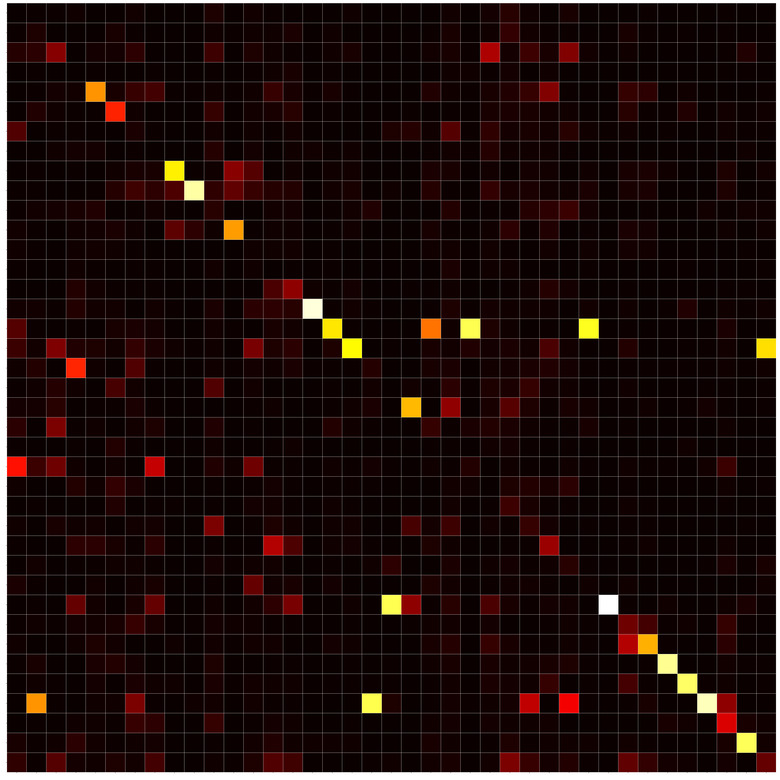}
         & 
         \includegraphics[height=5cm]{images/DCI/hot-colormap/colorbar.png}
    \end{tabular}

    \begin{tabular}{ccc}
        \LFCBM & \LABO &  \VLGCBM \\
         \includegraphics[width=0.3\textwidth]{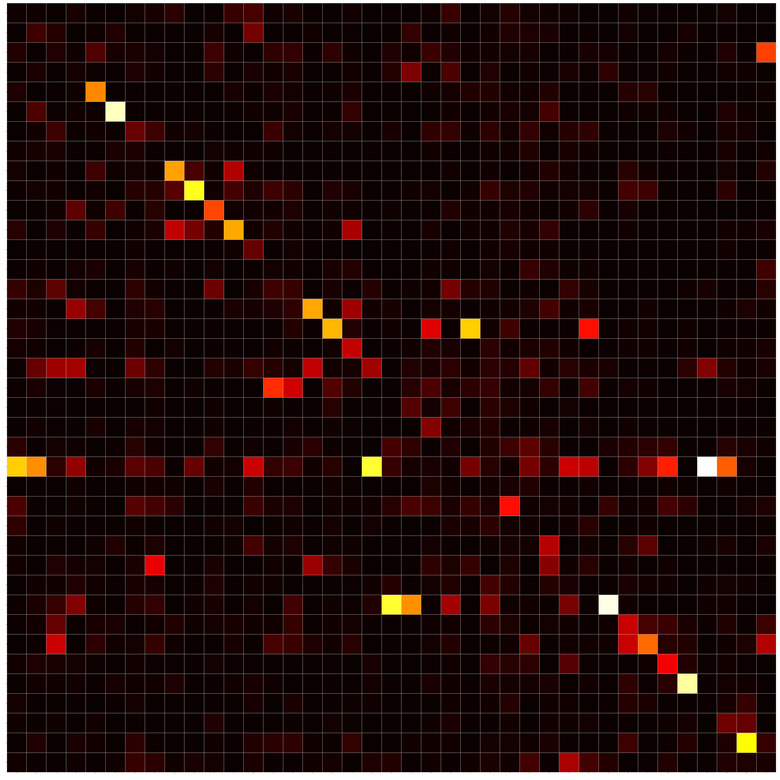}
         & 
         \includegraphics[width=0.3\textwidth]{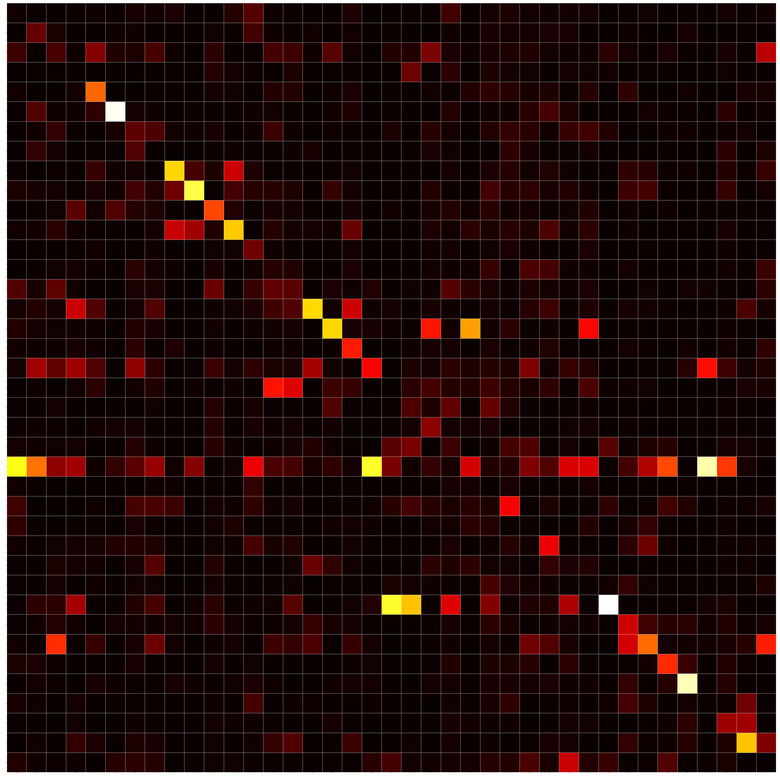}
         & 
         \includegraphics[width=0.3\textwidth]{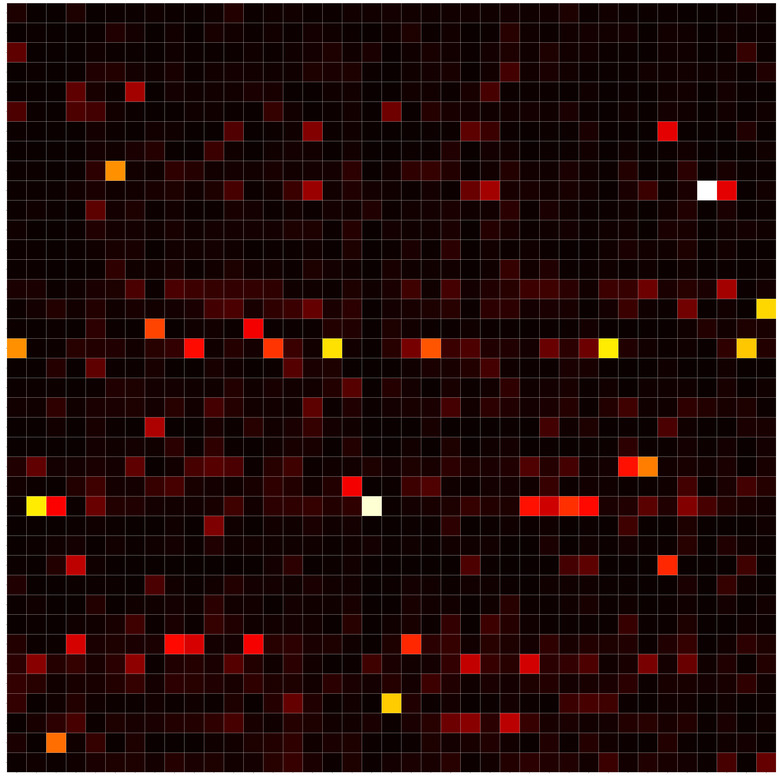}
    \end{tabular}
    \caption{\textbf{Importance matrices in DCI on \CelebA}. High disentanglement is achieved when, for each row, only one square is active. For reference, a perfect DCI matrix is a diagonal matrix.}
    \label{fig:disent-vlm-cbms-celeba}
\end{figure}

\begin{figure}[!t]
    \centering
    \begin{tabular}{ccr}
        \CBM & \method  \\
         \includegraphics[width=0.3\textwidth]{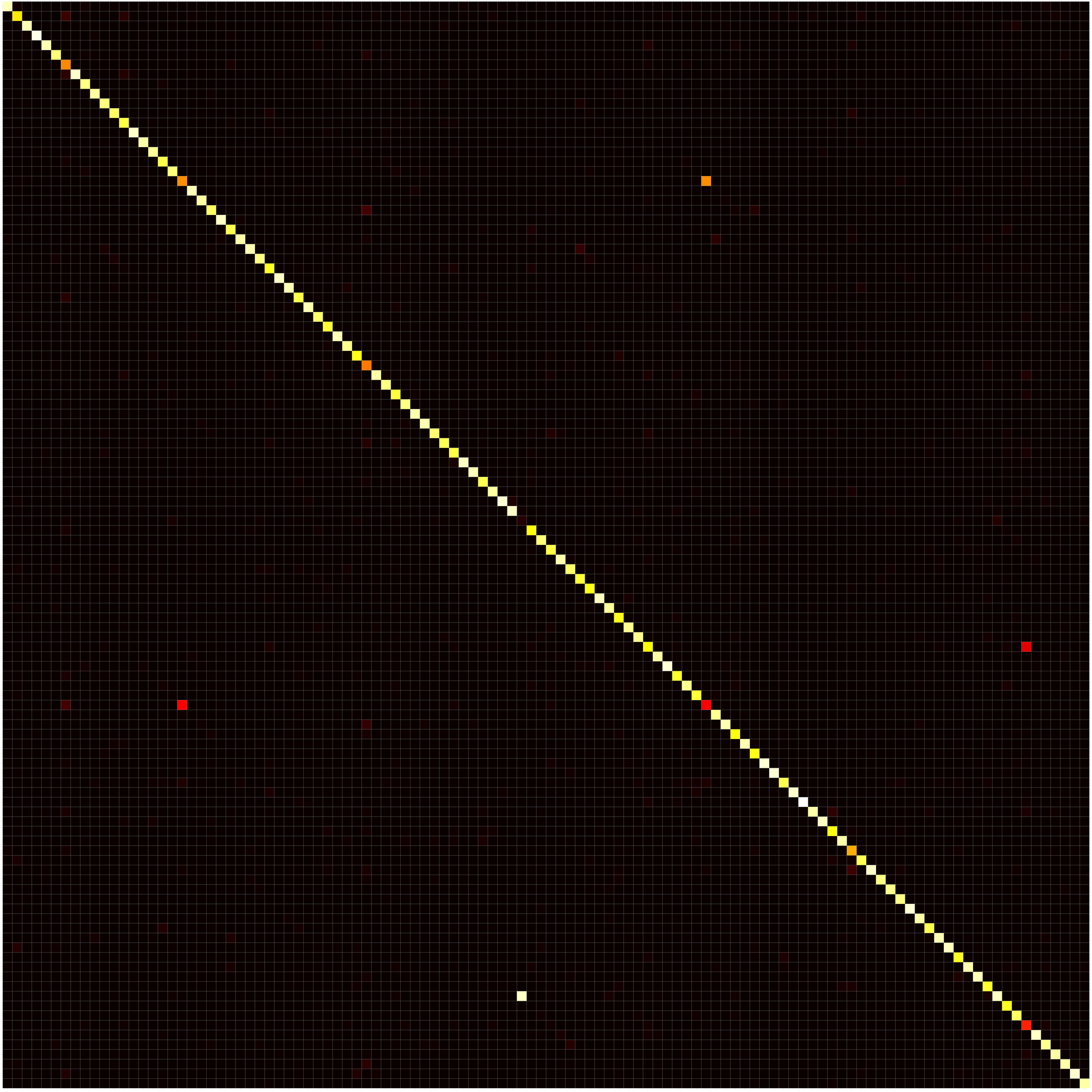}
         & 
         \includegraphics[width=0.3\textwidth]{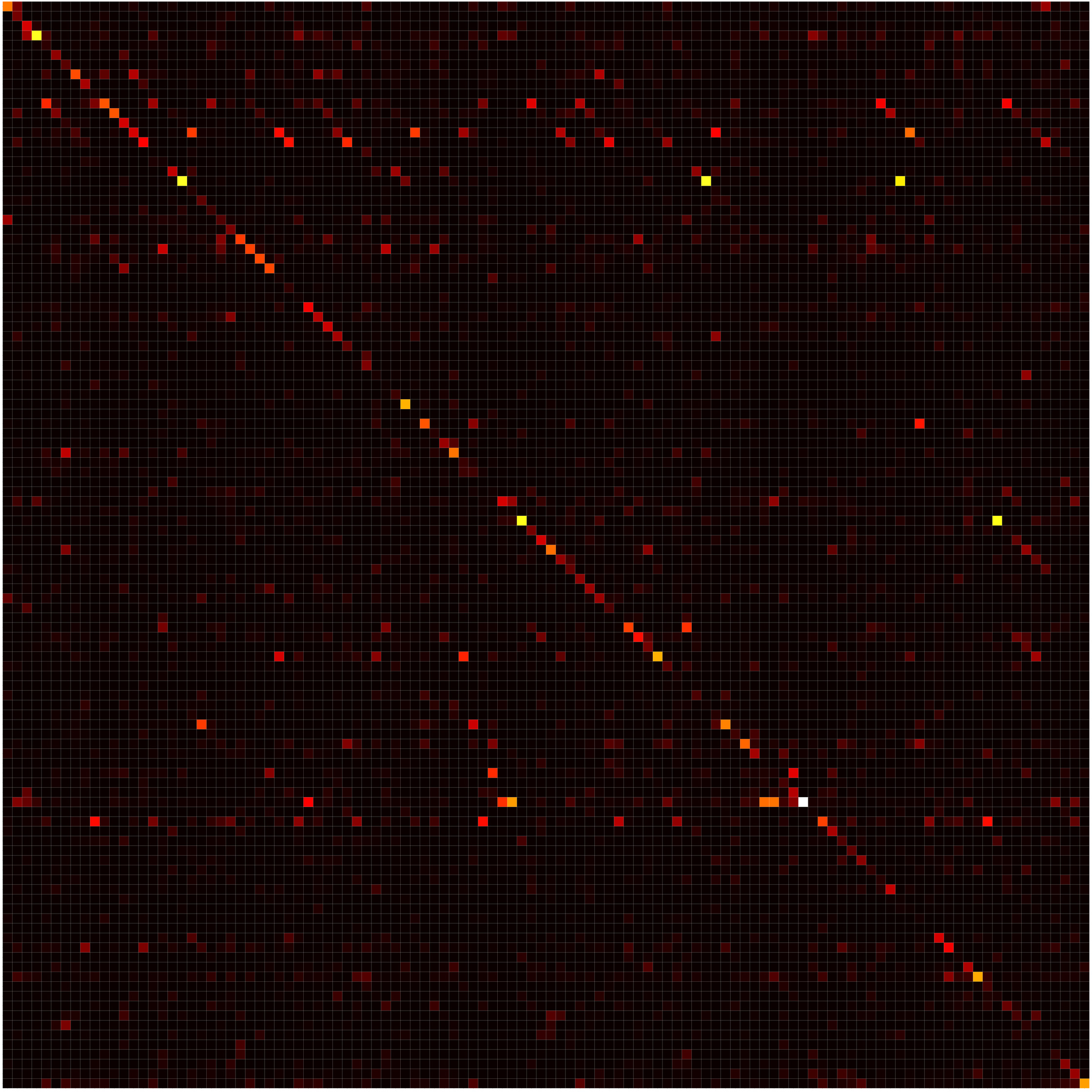}
         & 
         \includegraphics[height=5cm]{images/DCI/hot-colormap/colorbar.png}
    \end{tabular}

    \begin{tabular}{ccc}
        \LFCBM & \LABO &  \VLGCBM \\
         \includegraphics[width=0.3\textwidth]{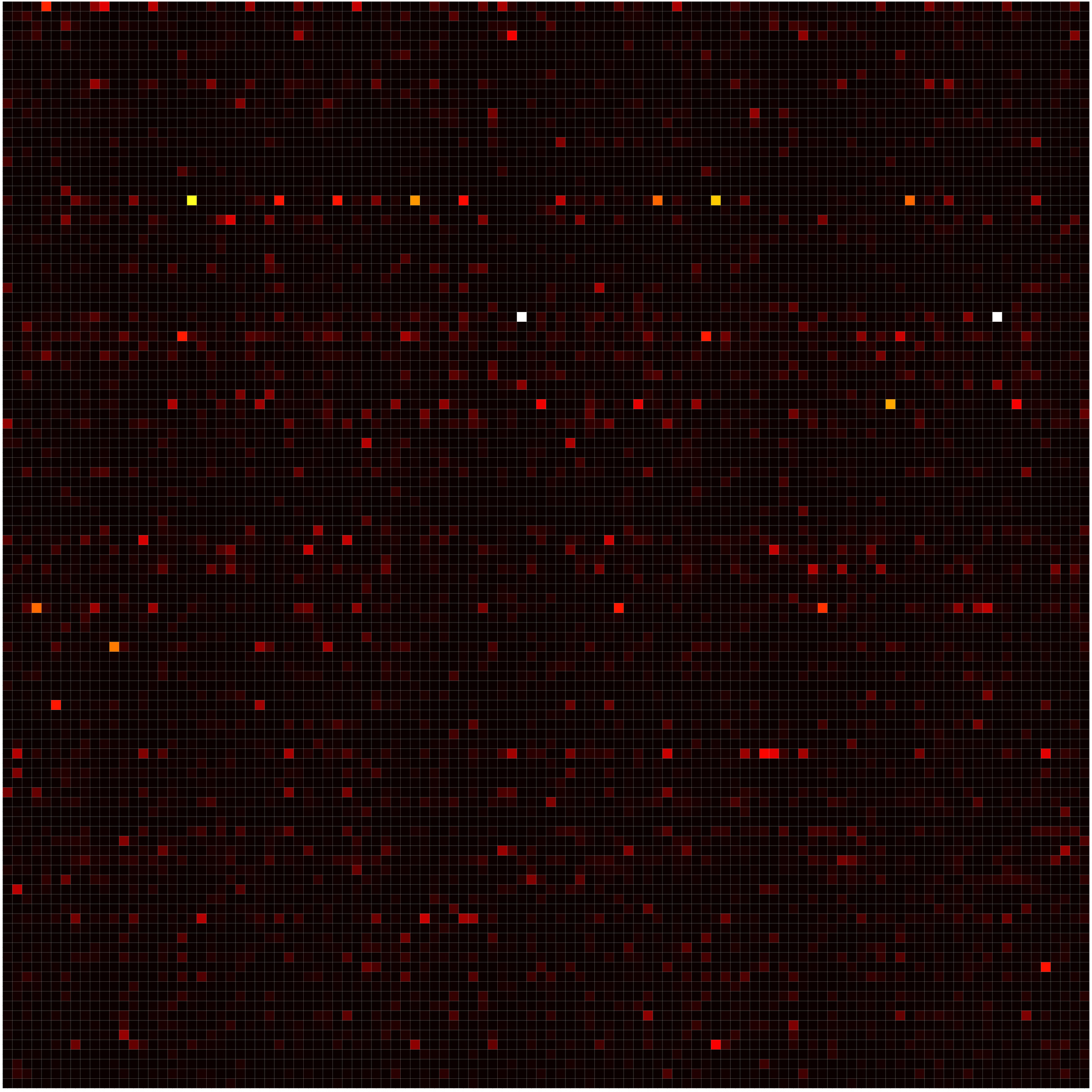}
         & 
         \includegraphics[width=0.3\textwidth]{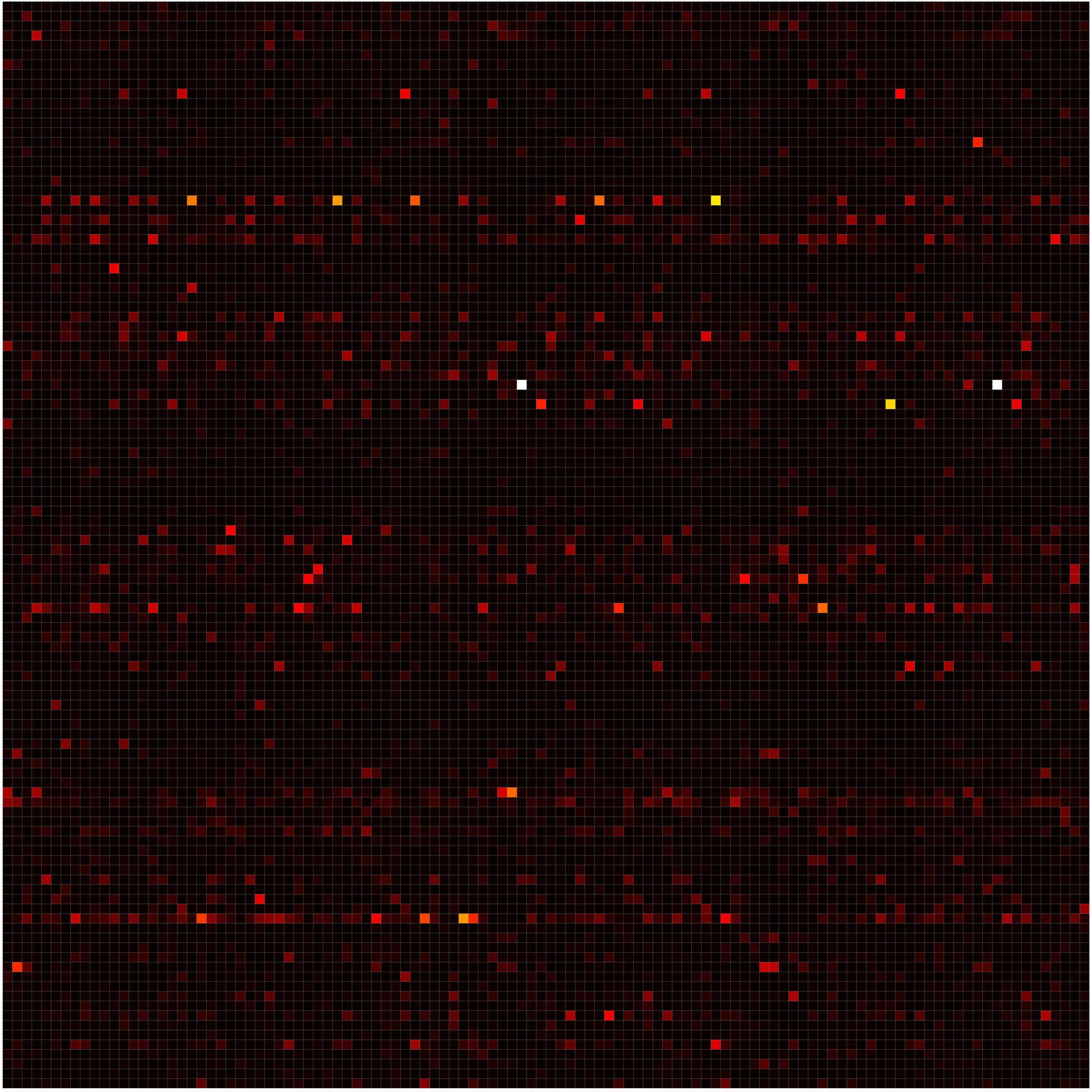}
         & 
         \includegraphics[width=0.3\textwidth]{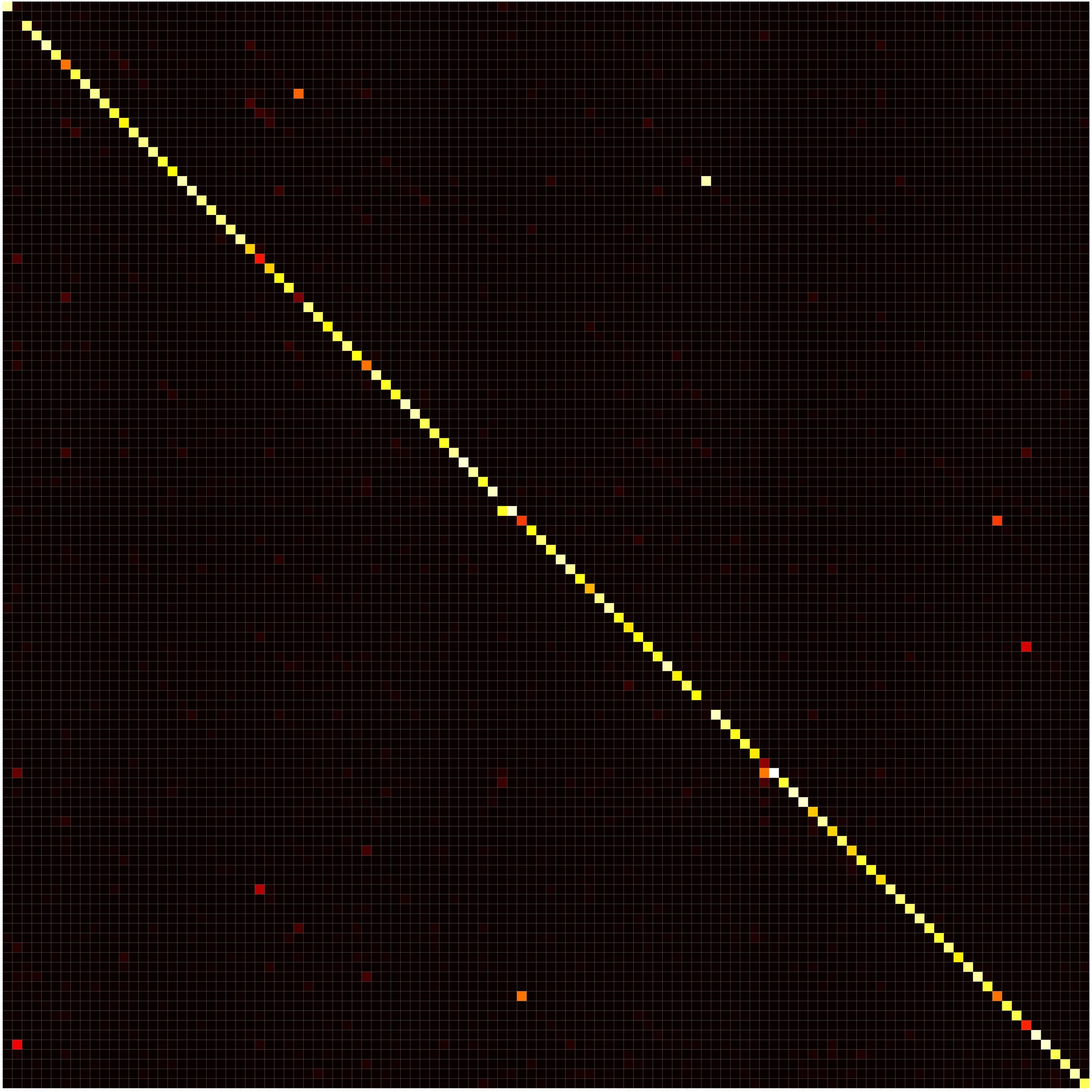}
    \end{tabular}
    \caption{\textbf{Importance matrices in DCI on \CUB}. High disentanglement is achieved when, for each row, only one square is active. For reference, a perfect DCI matrix is a diagonal matrix.}
    \label{fig:disent-vlm-cbms-cub}
\end{figure}

We provide in \cref{fig:disent-vlm-cbms-shapes3d}, \cref{fig:disent-vlm-cbms-celeba}, and \cref{fig:disent-vlm-cbms-cub} the DCI matrices obtained from the different models on \SHAPES, \CelebA, and \CUB, respectively.

\end{appendices}

\end{document}